\documentclass[preprint,12pt]{elsarticle}

\usepackage{amssymb}
\usepackage{amsmath}

\usepackage[utf8]{inputenc} 
\usepackage[T1]{fontenc}    
\usepackage{hyperref}      
\usepackage{url}           
\usepackage{booktabs}       
\usepackage{amsfonts}       
\usepackage{nicefrac}       
\usepackage{microtype}      
\usepackage{lipsum}		
\usepackage{graphicx}
\usepackage{natbib}
\usepackage{doi}
\usepackage{amsmath}
\usepackage{amssymb}
\usepackage{booktabs}
\usepackage{listings}
\usepackage{color}
\usepackage{algorithm} 
\usepackage{multirow}
\usepackage{colortbl}
\usepackage{listings}
\usepackage{makecell}
\usepackage{xcolor}
\usepackage{tabularray}
\usepackage[normalem]{ulem}
\usepackage{cuted}
\usepackage[inkscapelatex=false]{svg}
\usepackage{subfig}
\usepackage{subcaption}

\usepackage{soul}

\journal{Expert Systems With Applications}

\begin{document}

\newsavebox\CBox
\def\textBF#1{\sbox\CBox{#1}\resizebox{\wd\CBox}{\ht\CBox}{\textbf{#1}}}

\begin{frontmatter}



\title{Learning Motion Blur Robust Vision Transformers for Real-Time UAV Tracking}


\author[1]{You Wu}
\author[2]{Xucheng Wang}
\author[3]{Dan Zeng}
\author[1]{Hengzhou Ye}
\author[1]{Xiaolan Xie}
\author[4]{Qijun Zhao}
\author[1]{Shuiwang Li\corref{cor1}}
\ead{lishuiwang0721@163.com}
\cortext[cor1]{Corresponding author}

\affiliation[1]{organization={College of Computer Science and Engineering},
            addressline={Guilin University of Technology}, 
            city={Guilin},
            postcode={541004}, 
            country={China}}
\affiliation[2]{organization={School of Computer Science},
            addressline={Fudan University}, 
            city={Shanghai},
            postcode={200082}, 
            country={China}}
\affiliation[3]{organization={School of Artificial Intelligence},
            addressline={Sun Yat-sen University}, 
            city={Zhuhai},
            postcode={510275}, 
            country={China}}
\affiliation[4]{organization={College of Computer Science},
            addressline={Sichuan University}, 
            city={Chengdu},
            postcode={610065}, 
            country={China}}

\begin{abstract}
Unmanned aerial vehicle (UAV) tracking is critical for applications like surveillance, search-and-rescue, and autonomous navigation. However, the high-speed movement of UAVs and targets introduces unique challenges, including real-time processing demands and severe motion blur, which degrade the performance of existing generic trackers. While single-stream vision transformer (ViT) architectures have shown promise in visual tracking, their computational inefficiency and lack of UAV-specific optimizations limit their practicality in this domain.
In this paper, we boost the efficiency of this framework by tailoring it into an adaptive computation framework that dynamically exits Transformer blocks for real-time UAV tracking. 
The motivation behind this is that tracking tasks with fewer challenges can be adequately addressed using low-level feature representations. 
Simpler tasks can often be handled with less demanding, lower-level features. This approach allows the model use computational resources more efficiently by focusing on complex tasks and conserving resources for easier ones.
Another significant enhancement introduced in this paper is the improved effectiveness of ViTs in handling motion blur, a common issue in UAV tracking caused by the fast movements of either the UAV, the tracked objects, or both.
This is achieved by acquiring motion blur robust representations through enforcing invariance in the feature representation of the target with respect to simulated motion blur. We refer to our proposed approach as BDTrack. Extensive experiments conducted on four tracking benchmarks validate the effectiveness and versatility of our approach, demonstrating its potential as a practical and effective approach for real-time UAV tracking.
Code is released at: 
\url{https://github.com/wuyou3474/BDTrack}.
\end{abstract}






\end{frontmatter}


\section{Introduction}

In recent years, Artificial Intelligence (AI) has brought transformative progress across various domains, including AI-driven models~\cite{guo2024artificial,guo2025ai}, multimodal learning~\cite{liu2023visual,xu2023multimodal}, dynamic resource allocation~\cite{Tang2022YouNM,Xu2023LGViTDE}, unmanned aerial vehicle (UAV) tracking~\cite{cao2021hift,cao2022tctrack,li2023adaptive}, and intelligent energy systems and control applications~\cite{el2024artificial,fawzy2021modified,abulkhair2024negative,abdelsattar2024assessing,abdelsattar2024energy,abdelsattar2023overview,diab2019robust}.
Recent advancements have broadly improved system performance, energy efficiency, and real-time decision-making across various engineering fields. 
These developments have motivated us to apply them to UAV tracking, aiming to improve accuracy, enhance robustness in dynamic environments, and achieve efficient real-time processing.
UAV tracking involves the process of determining and predicting the location and size of a specific object in successive aerial images. This capability has become increasingly essential in a variety of fields, such as public safety, environmental monitoring, industrial inspections, disaster relief, and agriculture, due to its wide-ranging applications and benefits~\cite{cao2021hift,cao2021siamapn++,li2023adaptive,liu2023bactrack,yuan2024multi,li2024learning}.
These applications present unique challenges, including the need for accurate tracking in resource-constrained environments, where UAVs often face limited battery life and computational resources. Therefore, efficient tracking algorithms that ensure sustained and accurate performance under these conditions are paramount \cite{cao2022tctrack,li2022learning}.
Traditional discriminative correlation filter (DCF)-based trackers have been widely used in UAV tracking due to their efficiency \cite{henriques2015high,Huang2019LearningAR,li2020autotrack,lin2020learning}. However, their tracking precision often falls behind that of deep learning (DL)-based methods, which have shown promising results in terms of precision \cite{cao2021siamapn++,cao2022tctrack,liu2023bactrack,yuan2024multi}. Despite the advantages of DL-based methods, they face significant challenges in real-time applications, especially in resource-limited scenarios, where both computational power and battery life are constrained. Recent advances in the tracking community have led to the emergence of single-stream architectures, particularly those leveraging pre-trained vision transformer (ViT) backbone networks. These architectures have demonstrated notable success in integrating feature extraction and correlation \cite{ye2022joint,chen2022backbone,xie2022correlation,tang2023learning}. 
In line with this trend, Aba-ViTrack~\cite{li2023adaptive} introduces an efficient DL-based tracker that adopts a single-stream ViT framework with adaptive token selection, achieving a good balance between precision and speed. However, its variable token scheme incurs unstructured access overhead. Motivated by this, we adopt a similar architecture but enhance efficiency through a structured early exiting strategy.


\begin{figure*}[t]
\centering
\includegraphics[width=0.85\linewidth]{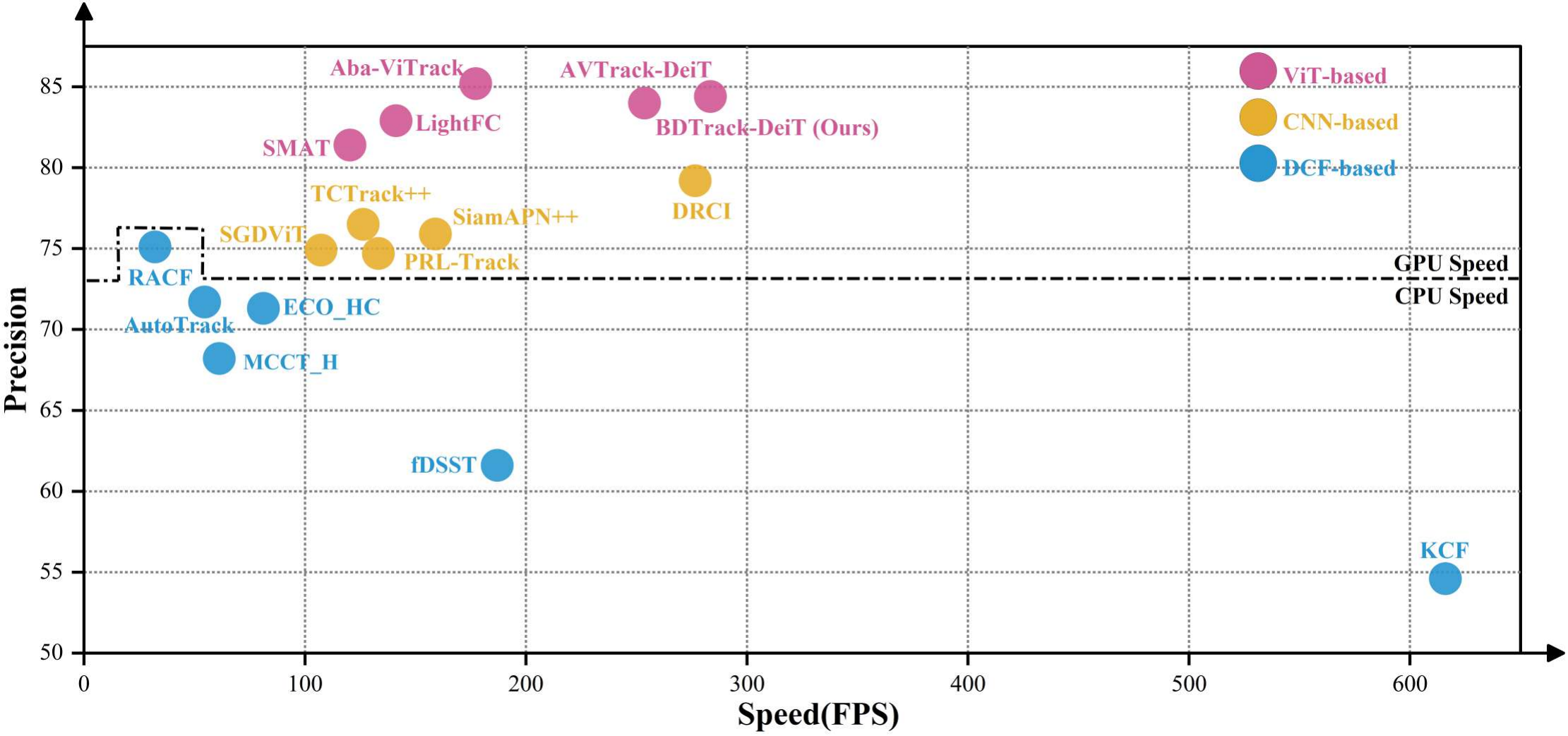}
\caption{Compared with state-of-the-art UAV tracking algorithms on four UAV tracking benchmarks, our BDTrack-DeiT sets a new record with 0.844 precision while maintaining efficient performance at around 283 FPS.}\label{fig_prec_fps}
\end{figure*}

Early exiting is a technique designed to expedite the processing of input data by allowing a model to terminate its forward pass prematurely based on certain criteria, reducing computational costs without compromising much predictive accuracy, especially if the behavior or characteristics of the task vary based on the specific examples or instances involved. 
While previous research has mainly explored this idea in domains such as natural language processing and image classification \cite{Kaya2018ShallowDeepNU,Li2019ImprovedTF,Liu2020FasterDT,Xin2021BERxiTEE,Xu2023LGViTDE,zhou2020bert}, visual tracking, particularly real-time UAV tracking, poses unique challenges that differentiate it from these static tasks. Unlike image classification, visual tracking requires maintaining temporal consistency across frames, coping with abrupt target motion, deformation, scale variation, and appearance changes. 
These properties demand early-exit strategies that not only assess the model’s confidence for a single frame but also account for the temporal dependency of predictions and the evolving target appearance over time. Existing studies have investigated diverse strategies, including adaptive confidence-based exits \cite{Kaya2018ShallowDeepNU,Xu2023LGViTDE}, distillation \cite{Phuong2019DistillationBasedTF}, architectural design \cite{Teerapittayanon2016BranchyNetFI,Liu2020FasterDT}, and specialized training schemes \cite{Li2019ImprovedTF,Xin2021BERxiTEE}, applied to architectures ranging from convolutional neural networks (CNNs) \cite{Kaya2018ShallowDeepNU,Phuong2019DistillationBasedTF} to recurrent neural networks (RNNs) \cite{Teerapittayanon2016BranchyNetFI} and Transformers \cite{Liu2020FasterDT,Xu2023LGViTDE}. While these methods have shown success in applications such as NLP \cite{Teerapittayanon2016BranchyNetFI,Xin2021BERxiTEE}, image classification \cite{Xu2023LGViTDE}, video recognition \cite{Ghodrati2021FrameExitCE}, image captioning \cite{Fei2022DeeCapDE}, and vision-language learning \cite{Tang2022YouNM}, their direct application to UAV tracking remains largely unexplored. This gap motivates our work, which adapts early-exit mechanisms to the real-time, temporally dependent, and appearance-variant nature of UAV tracking.

In addition, this research also aims to enhance the performance of UAV trackers by mitigating the adverse effects of motion blur, which is a prevalent challenge in UAV tracking as rapid and unpredictable movements of UAVs are common due to factors such as wind, changes in altitude, sudden maneuvers, or the need to track fast-moving targets \cite{li2020autotrack,li2022learning,Fu2020CorrelationFF}. 
For example, existing trackers ~\cite{cao2021siamapn++,cao2022tctrack} show a marked increase in center location error (CLE) during real-world testing with motion-blurred sequences, rising from under 10 pixels to over 20, a threshold commonly regarded as tracking failure. This degradation becomes especially severe during aggressive UAV maneuvers such as sharp turns or rapid accelerations, where motion blur is significantly intensified, further impairing tracking performance.
Although numerous methods have been proposed to address motion blur in visual tracking~\cite{ma2016visual,seibold2017model,guo2021learning,guo2019effects,zuo2023adversarial}, their practical use in real-time UAV tracking remains limited due to inherent shortcomings.
In contrast, we propose to learn motion blur robust feature representations with ViTs by minimizing the mean square error (MSE) between the feature representation of the original template and its blurred version subject to motion blur.
Our approach distinguishes itself through its simplicity, avoiding the inclusion of extra architectural complexities.
The proposed framework is called BDTrack. 
Validation using real-world data demonstrates that our method significantly enhances the robustness against motion blur of baseline methods. Remarkably, our BDTrack-DeiT outperforms the SOTA tracker ABDNet \cite{zuo2023adversarial}, which was proposed to address motion blur, by 5.5\% in precision on the motion blur subset of the UAVDT dataset with more than 2 times faster GPU speed (See Fig. \ref{fig_attr_plots} and Table \ref{table_lightweight}). Extensive experiments on four benchmarks show that our method achieves SOTA performance.
As shown in Fig. \ref{fig_prec_fps}, our method sets a new record with a average precision of 0.844 and runs efficiently at around 283 frames per second (FPS) on the four UAV tracking benchmarks.
Its real-time, motion-blur-robust tracking improves UAV efficiency in applications such as agricultural inspection by enabling faster and wider-area coverage with minimal computational load, and in disaster relief operations by maintaining reliable tracking in challenging conditions to support quicker and more effective search-and-rescue missions. We also emphasize its suitability for resource-constrained UAVs through adaptive computation that preserves battery life and processing power.

The following is a summary of our contributions:

\begin{itemize}

    \item We introduce a novel module that adaptively streamlines the architecture of ViTs by dynamically exiting Transformer blocks. Our method is straightforward and can significantly enhance the efficiency of baseline ViT-based UAV tracking methods with minimal impact on their tracking performance.

    \item We make a pioneering effort to enhance the motion blur robustness of vision transformers (ViTs) for real-time UAV tracking by leveraging a simple mean square error (MSE) loss to enforce feature invariance of the target under simulated blur. Our approach incurs no additional computational burden during the inference, making it well-suited for real-time UAV tracking.

    \item We present BDTrack, an efficient tracker seamlessly integrating these components, which is readily integrable into similar ViT-based trackers. BDTrack is able to maintain high tracking speeds while delivering exceptional performance. Extensive experiments on four benchmarks confirm its state-of-the-art performance.

\end{itemize}


\section{Related Work}\label{section_related_work}

\subsection{Motion Blur Aware Trackers}
Motion blur presents a significant challenge in visual tracking, and various approaches have been proposed to address this issue and enhance the robustness of tracking algorithms \cite{ding2015severely,seibold2017model,guo2021learning,guo2019effects,zuo2023adversarial}, including employing specialized architectures or mechanisms to handle motion blur, learning motion blur robust representations, and combining motion deblurring techniques. 
For example, 
Ding et al. \cite{ding2015severely} introduce a method to enhance the robustness of the appearance model against severe blur by incorporating various blur kernels during the training process.
Guo et al. \cite{guo2019effects} propose a novel generative adversarial network (GAN)-based scheme aimed at enhancing tracker robustness to motion blurs. To address the challenges posed by motion blur in UAV tracking, 
Zuo et al. \cite{zuo2023adversarial} introduced the tracking-oriented adversarial blur-deblur network (ABDNet) very recently, which includes a novel deblurring component to restore the visual appearance of blurred targets and an innovative blur generator that creates realistic blurry images for adversarial training.
However, existing methods for handling motion blur often involve complex architectures \cite{mao2021robust,zuo2023adversarial} or training pipelines \cite{guo2021learning,zuo2023adversarial} and neglect runtime efficiency \cite{ma2016visual,seibold2017model,guo2019effects},limiting their suitability for UAV applications. 
We introduce motion blur robust ViTs for UAV tracking that learn invariant feature representations efficiently without adding inference overhead, making them ideal for real-time scenarios. The method is easily adaptable to other ViT-based trackers, improving performance under motion blur without sacrificing speed.

\subsection{Efficient Vision Transformers}
Efforts to enhance the efficiency of ViTs have garnered significant attention in recent years, driven by the need to reconcile their powerful representation capabilities with computational efficiency. One avenue of exploration involves the design of lightweight ViT architectures, employing techniques such as low-rank methods, model compression, and hybrid designs~\cite{Yang2023SkeletonNN,Mao2021TPruneET,Li2022EfficientFormerVT}.
However, low-rank and quantization-based ViTs often trade substantial accuracy for efficiency gains. Pruning-based ViTs typically require intricate decisions on pruning ratios and involve a time-consuming fine-tuning process. Hybrid ViTs featuring CNN-based stems impose limitations on input size flexibility, preventing the simultaneous input of images with varying dimensions \cite{li2023adaptive}.


Conditional computation-based ViTs have recently enabled adaptive inference by dynamically adjusting computation based on input complexity~\cite{Rao2021DynamicViTEV,yin2022vit}. For example, A-ViT~\cite{yin2022vit} eliminates extra halting networks via ACT-like mechanisms, improving both efficiency and accuracy. 
Aba-ViTrack~\cite{li2023adaptive} extends this design to UAV tracking, but suffers from the overhead of unstructured token selection. In contrast, our DEEM adopts a structured early-exit strategy that enables efficient inference, implemented using only a simple linear layer with sigmoid activation.
Nevertheless, existing studies primarily focus on large-scale models and incorporate inefficient architectures or training methods for making early exiting decisions~\cite{Xin2021BERxiTEE,Fei2022DeeCapDE,Tang2022YouNM,Xu2023LGViTDE}. These criteria for early exiting are challenging to adapt effectively to lightweight models.
For example, the early exiting strategy proposed for vision language models in \cite{Tang2022YouNM} relies on layer-wise cosine similarities.
While cosine similarities have proven effective for large-scale models characterized by high-level and high-dimensional feature representations, their performance may be suboptimal in lightweight models where non-linearities and complex dependencies are challenging to linearize or trivialize \cite{Sivertsen2019SimilarityPI,Srivamsi2023CosineSB}. 
The dynamic early exiting method proposed for accelerating ViTs in \cite{Xu2023LGViTDE} involves training delicate Local Perception and Global Aggregation Heads separately with distinct supervisions. However, this intricate training setup introduces significant inefficiencies into the training process.
In this work, we explore a more structured and efficient approach to dynamically early exit ViT blocks by incorporating a dynamic early exit module.
Our method aims to simplify the training process, improve inference efficiency, and maintain adaptability across different ViT-based models.

\begin{figure*}[t]
	\centering
\includegraphics[width=1\textwidth]{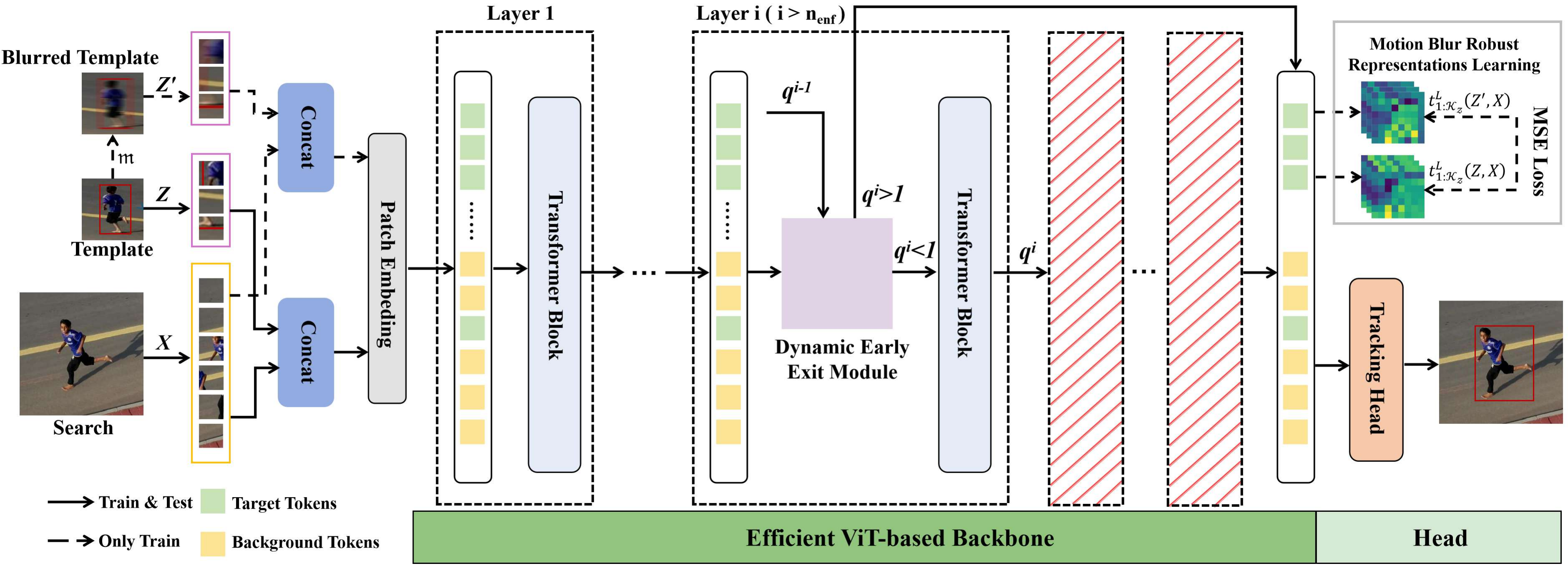}
\caption{Overview of the proposed BDTrack. It comprises of a single-stream backbone and a prediction head. MBRV and DEEM (see Fig. \ref{BDTrack-DDEM}) are employed to learn motion blur robust representations and to boost efficiency, respectively. 
Note that $\mathfrak{m}$ applies simulated motion blur to the template image \( Z \) using the method from \cite{Dwibedi2017CutPA}. DEEM takes a subset of tokens from the target template and search image as input, outputting the exit score (i.e., $q$) at the corresponding ViT block.}
 \label{BDTrack}
\end{figure*}

\section{Method}\label{section_method}
In this section, we first briefly overview our tracking framework, BDTrack, as illustrated in Fig. \ref{BDTrack}. Then, we introduce the proposed method for learning motion blur robust ViTs (MBRV) and the dynamic early exit module (DEEM). Finally, we detail the prediction head and training objective.

\subsection{Overview}
Our BDTrack is a tracking framework that employs a single-stream approach, taking the target template and the search image as inputs. This framework consists of an adaptive ViT-based backbone and a prediction head. The features extracted from the backbone are subsequently forwarded to the prediction head to achieve tracking predictions.
To enhance the robustness of ViTs against motion blur, we minimize the mean squared error (MSE) between the feature representations of the target template and its blurred counterpart. The blurred version of the template is generated using the linear motion blur method, as referenced in \cite{MethodsinMedicine2023RetractedUA}. This approach aims to ensure that the ViTs can maintain consistent performance even when the input images are affected by motion blur, a common issue in UAV tracking.
Additionally, to further enhance the efficiency of ViTs, we introduce a dynamic early exit module (DEEM) for each ViT block (layer), excluding the initial $n_{enf}$ blocks.
This module is trained to dynamically predicts the exiting score $\mathfrak{e}^l$ at the $l$-th layer. The forward pass of the ViT exists at layer $l$ when the cumulative exiting score exceeds a certain threshold. 
This mechanism allows the model to allocate computational resources more efficiently, exiting early for simpler tasks and reserving full processing power for more complex ones.
The details of the two components will be elaborated in the subsequent subsections.

\subsection{Learning Motion Blur Robust ViTs (MBRV)}
A motion blur robust ViT (MBRV) serves as a pivotal component in our pursuit of a compact and efficient end-to-end tracker specifically designed for real-time UAV tracking. In the context of UAV tracking, where rapid and accurate object localization is paramount, the presence of motion blur in captured images poses a significant challenge. Motion blur can distort object features, leading to inaccuracies in tracking algorithms and potentially compromising the effectiveness of the entire tracking system. The inclusion of MBRV addresses these challenges by ensuring that the ViT-based backbone of our tracking system remains effective even in the presence of motion blur. By minimizing the Mean Squared Error (MSE) between the feature representations of the target template and its blurred counterpart, MBRV enhances the robustness of the tracker. Furthermore, the compact and efficient nature of MBRV aligns with our goal of developing a tracker optimized for real-time UAV tracking. 
It is worthy of note that we choose to blur the template patch rather than the search patch is based on the following considerations: 1) It allows a controllable blurring of the target, as the object in the template always is sharp and unblurred while the object in the search patch is frequently subject to motion blur. 2) Since the size of the template patch and the scale of the target are fixed, the tokens associated with the template can be precisely identified. But the identification of the objects in the search patch may involve interpolation due to translation and scale variations of the objects. As the size of the search patch is fixed, interpolation has to be conducted on the feature space. However, interpolating in the feature space can lead to several issues, including loss of spatial information, misalignment of features, and inconsistent semantic meaning. These factors can result in poorer quality interpolated features, affect training dynamics, and potentially lead to suboptimal model performance.

During the training phase, the MBRV receives three inputs: a target template $Z$, a blurred target template $Z^{\prime}$= $\mathfrak{m}(Z)$, and a search image $X$, where
$\mathfrak{m}(Z)$ represents applying simulated motion blur to the template image $Z$ using the linear motion blur method implemented in \cite{Dwibedi2017CutPA}.
The input images are first segmented and then flattened into sequences of patches. Subsequently, the sequences corresponding to $Z$ and $X$ are concatenated into an ordered sequence denoted by $[Z, X]$. The same procedure is applied to the sequences corresponding to $Z'$ and $X$, leading to $[Z', X]$.  In the forward pass, $[Z, X]$ is tokenized by a trainable linear projection layer $\mathcal{E}$ called patch embedding, which generates $\mathcal{K}$ tokens as follows:
\begin{equation}
\textbf{t}_{1:\mathcal{K}}^0=\mathcal{E}([Z,X])\in \mathbb{R}^{\mathcal{K}\times d},
\end{equation}
where the embedding dimension of each token is denoted by $d$, and the token sequences $t^0_{1:\mathcal{K}z}$ and $t^0_{\mathcal{K}_z+1:\mathcal{K}}$ ($1<\mathcal{K}_z<\mathcal{K}_x<\mathcal{K}$) correspond to the template and search image, respectively, where $\mathcal{K}=\mathcal{K}_z+\mathcal{K}_x$. The $\mathcal{K}$ tokens $\textbf{t}_{1:\mathcal{K}}^0$ are then input into the Transformer blocks to obtain their final feature representations.
Let $\mathfrak{T}^l$ represent the Transformer block at layer $l$, transforming all tokens from layer $(l-1)$ as $\textbf{t}_{1:\mathcal{K}}^l=\mathfrak{T}^l(\textbf{t}_{1:\mathcal{K}}^{l-1})$. The overall MBRV of $L$ ViT blocks in total, denoted by $\mathfrak{B}$, can be expressed as:
\begin{equation}	
\textbf{t}_{1:\mathcal{K}}^L(Z,X)=\mathfrak{B}(Z,X)=\mathfrak{T}^L\circ \mathfrak{T}^{L-1} \circ ...\circ \mathfrak{T}^1 \circ \mathcal{E}([Z,X]) ,
\end{equation}
where $\circ$ denotes the composition operation. Similarly, we can derive the feature representation of $[Z',X]$.
As $[Z,X]$ and $[Z',X]$ are ordered sequences of tokens, the feature representation of $Z$ and $Z'$ can be retrieved by tracking their token indices in the respective ordered sequences, specifically denoted as $\textbf{t}_{1:\mathcal{K}z}^L(Z,X)$ and $\textbf{t}_{1:\mathcal{K}_z}^L(Z',X)$, respectively.
The core idea behind MBRV is to minimize the mean square error between the feature representation of $Z$ and that of $Z'$. This is achieved by minimizing the following MSE loss:
\begin{equation}\label{Eq_MI_loss}	\mathcal{L}_{br}=||\textbf{t}_{1:\mathcal{K}_z}^L(Z,X)-\textbf{t}_{1:\mathcal{K}_z}^L(Z',X)||^2.
\end{equation}
During the inference phase, only the sequence $[Z,X]$ is input to the MBRV, there is no need for simulating motion blur. Consequently, our method does not introduce any additional computational cost in the inference phase. Notably, our method is agnostic to the specific ViTs employed, allowing seamless integration with any ViTs within our framework.

\begin{figure}[h]
	\centering
\includegraphics[width=0.48\textwidth]{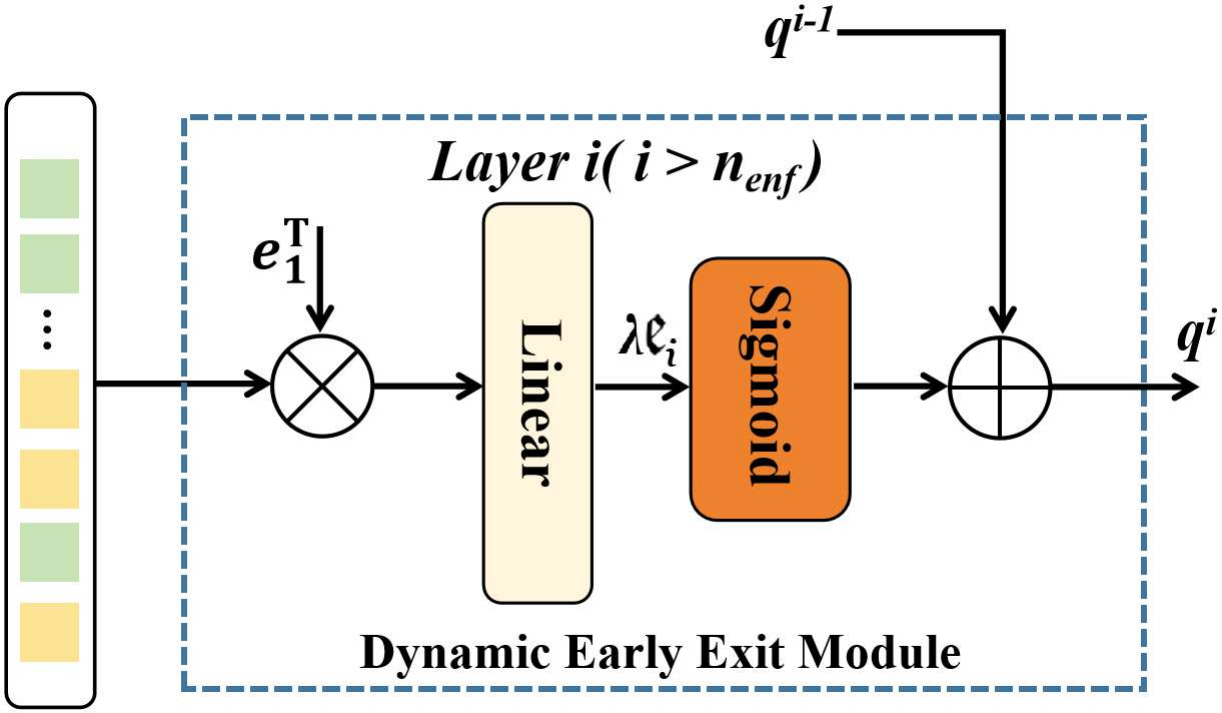}
\caption{The detailed structure of the DEEM. The variables $\textbf{e}_1^{\textup{T}}$ represents a subset of all tokens output by the \((i-1)\)-th ViT block, $\mathfrak{e}_i$ denotes the exit score of the \(i-\)th Transformer block, Sigmoid represents the sigmoid function, and $q^i$ denotes the cumulative exit score before $(i-1)$.}
 \label{BDTrack-DDEM}
 \vspace{-0.3cm}
\end{figure}

\subsection{Dynamic Early Exit Module (DEEM)}

The primary goal of the DEEM is to allocate computational resources in a dynamic manner based on the complexity of each VIT block's input, aiming to enhance inference efficiency while preserving performance. The DEEM operates by dynamically assessing the complexity of the input at each ViT block and making decisions regarding whether to continue processing or to exit early. This decision-making process is based on predefined criteria, specifically the cumulative exiting score exceeding a certain threshold. By exiting early for simpler inputs and reserving computational resources for more complex ones, the module optimizes the overall efficiency of the tracking system.
While the primary objective of the DEEM is to improve inference efficiency, it achieves this with minimal impact on tracking performance.
The module is designed to ensure that early exits occur only when appropriate, thereby maintaining the accuracy and reliability of the tracking predictions. This balance between efficiency and performance is essential in real-time UAV tracking systems.

The DEEM is accomplished by employing a linear layer in conjunction with a non-linear activation function, with a slice of all tokens representing both the target template and the search image as its input, as illustrated in Fig. \ref{BDTrack-DDEM}.
Its output represents the exiting score at the corresponding ViT block (layer).
Specifically, take the $i$-th ($i>n_{enf}$) layer for example, the
slice of all tokens ouput by the $(i-1)$-th ViT block is denoted by
$\textbf{e}_1^{\textup{T}}\textbf{t}_{1:\mathcal{K}}^{i-1}(Z,X):=\textbf{b}^{i-1}\in \mathbb{R}^{\mathcal{K}}$, where $\textbf{e}_1^{\textup{T}}=[1,0,...,0]\in \mathbb{R}^{\mathcal{K}}$ is a standard unit vector in $\mathbb{R}^{\mathcal{K}}$,
the linear layer is denoted by $\mathfrak{l}^i$, where $d$ is the token dimension. 
Formally, the DEEM at layer $i$ is defined by $\mathfrak{e}^i=\sigma(\mathfrak{l}^i(\textbf{b}^{i-1}))$, 
where $\mathfrak{e}^i\in [0,1]$ represents the exiting score of the $i$-th Transformer block, $\sigma(x)=1/(1+e^{-x})$ denotes the sigmoid function. 
The forward pass exits if the cumulative exiting score $q^i=\sum_{k=n_{enf}+1}^{i}\lambda \mathfrak{e}^k$ exceeds $1-\epsilon$, i.e., $q^i\geqslant 1-\epsilon$, where $\epsilon$ is a small positive constant that enables early exiting after examining a single DEEM module. In our implementation, we set $\epsilon = 0.05$, which allows an early exit when the cumulative confidence exceeds 0.95. Note that, if $n_{enf}=0$, theoretically all $L$ blocks could be skipped simultaneously, resulting in no correlation computation between the template and search image. To prevent such an 
undesirable case, the first $n_{enf}>0$ layers are enforced without examination. This strategy is implemented to reduce computational burdens because these low-level blocks are expected in most cases, as low-level features provide foundational information upon which more complex features and representations can be built.
Another extreme case is that, for any input, early exit never happens as by which the model is easier to reduce the training losses with more ViT blocks. To address this, we introduce a block sparsity loss, denoted as $\mathcal{L}_{spar}$, which penalizes smaller average exiting scores over all examined layers. This encourages, on average, an earlier exit during the forward pass to enhance efficiency. Let $L_e$ be the maximum layer of examined ViT blocks, it follows that 
\begin{equation}
    L_e=\underset{n_{enf}< n\leqslant L}{\textup{argmin}}\left \{ \sum_{l=n_{enf}+1}^{n}\mathfrak{e}^l\geqslant 1-\epsilon  \right \} .
\end{equation}
The block sparsity loss is formally defined as follows:
\begin{equation}
\begin{split}
    \mathcal{L}_{spar}=||
    \frac{1}{L_e-n_{enf}}\sum^{L_e}_{l=n_{enf}+1}\mathfrak{e}^l-\tau||^2,
\end{split}
\end{equation}
where $\tau$ is a constant used along with $\lambda$ to control the block sparsity. We call $\tau$ the block sparsity constant. Given $\lambda$, generally the larger $\tau$ is, the more sparse the model will be.

\subsection{Prediction Head and Training Objective}
Following \cite{ye2022joint}, we utilize a fully convolutional network-based prediction head $\mathcal{C}$, consisting of multiple Conv-BN-ReLU layers.
The output tokens $t^{L_e}_{\mathcal{K}_z+1:\mathcal{K}}$ associated with the search image are first reinterpreted into a 2D spatial feature map, which is then input to the prediction head. The head generates three outputs: a target classification score $\mathbf{p} \in [0,1]^{H_x/P\times W_x/P}$, a local offset $\mathbf{o}\in [0,1]^{2\times H_x/P\times W_x/P}$, and a normalized bounding box size $\mathbf{s} \in [0,1]^{2\times H_x/P\times W_x/P}$.
The highest classification score is used to estimate the coarse target position, i.e., $(x_c, y_c)=\textup{argmax}_{(x,y)}\mathbf{p}(x,y)$, based on which the final target bounding box is determined by
\begin{equation}
		\{(x_t,y_t);(w,h)\}=\{(x_c, y_c)+\mathbf{o}(x_c,y_c);\mathbf{s}(x_c,y_c)\}.
\end{equation}
For the tracking task, we employ the weighted focal loss \cite{Law2018CornerNetDO} for classification and a combination of $L_1$ loss and GIoU loss \cite{Rezatofighi2019GeneralizedIO} for bounding box regression.
The overall loss function is formulated as:
\begin{equation}
\label{eq4}
 \mathcal{L}_{overall} =  \mathcal{L}_{cls} + \eta_{iou} \mathcal{L}_{iou} + \eta_{L_1} \mathcal{L}_{L1} + \rho \mathcal{L}_{br} + \gamma \mathcal{L}_{spar} 
\end{equation}
where the constants $\eta_{iou}=2$ and $\eta_{L_1}=5$ are the same as in \cite{ye2022joint}, $\rho$ and $\gamma$ are set to $10^{-4}$ and $10^{3}$ respectively. Our framework is trained end-to-end with the overall loss $\mathcal{L}_{overall}$ after the pretrained weights of the ViT for image classification is loaded.

\begin{table}[t]
\scriptsize
\centering
\setlength\tabcolsep{2pt}
\caption{
Comparison of precision (Prec.) and success rate (Succ.) between BDTrack and lightweight trackers on four UAV tracking benchmarks, namely UAVDT~\cite{du2018unmanned}, VisDrone2018~\cite{wen2018visdrone}, UAV123~\cite{mueller2016benchmark}, and UAV123@10fps~\cite{mueller2016benchmark}. 
{\color[HTML]{FE0000}Red}, {\color[HTML]{3531FF}blue} and {\color[HTML]{ 1f821c}green} indicate the first, second and third place. It is worth noting that the percent symbol (\%) has been omitted for Prec. and Succ. values.}
\label{table_lightweight}
\begin{tabular}{ccccccccccccc}
\toprule[1pt]
\multicolumn{2}{c}{}                         &                          & \multicolumn{2}{c}{UAVDT}                                                   & \multicolumn{2}{c}{VisDrone2018}                                                & \multicolumn{2}{c}{UAV123}                                                  & \multicolumn{2}{c}{UAV123@10fps}                                            & \multicolumn{2}{c}{Avg.}                                                    \\
\multicolumn{2}{c}{\multirow{-2}{*}{Method}} & \multirow{-2}{*}{Source} & Prec.                                & Succ.                                & Prec.                                & Succ.                                & Prec.                                & Succ.                                & Prec.                                & Succ.                                & Prec.                                & Succ.                                \\ \hline
                             & KCF~\cite{henriques2015high}            & TPAMI 15                 & 57.1                                 & 29.0                                 & 68.5                                 & 41.3                                 & 52.3                                 & 33.1                                 & 40.6                                 & 26.5                                 & 54.6                                 & 32.5                                 \\
                             & fDSST~\cite{danelljan2017discriminative}         & TPAMI 17                 & 66.6                                 & 38.3                                 & 69.8                                 & 51.0                                 & 58.3                                 & 40.5                                 & 51.6                                 & 37.9                                 & 61.6                                 & 41.9                                 \\
                             & BACF~\cite{kiani2017learning}          & ICCV 17                  & 68.6                                 & 43.2                                 & 77.4                                 & 56.7                                 & 66.0                                 & 45.9                                 & 57.2                                 & 41.3                                 & 67.3                                 & 46.8                                 \\
                             & ECO\_HC~\cite{Danelljan2017ECOEC}       & CVPR 17                  & 69.4                                 & 41.6                                 & 80.8                                 & 58.1                                 & 71.0                                 & 49.6                                 & 64.0                                 & 46.8                                 & 71.3                                 & 49.0                                 \\
                             & MCCT\_H~\cite{wang2018multi}       & CVPR 18                  & 66.8                                 & 40.2                                 & 80.3                                 & 56.7                                 & 65.9                                 & 45.7                                 & 59.6                                 & 43.4                                 & 68.2                                 & 46.5                                 \\
                             & ARCF~\cite{Huang2019LearningAR}          & ICCV 19                  & 72.0                                 & 45.8                                 & 79.7                                 & 58.4                                 & 67.1                                 & 46.8                                 & 66.6                                 & 47.3                                 & 71.4                                 & 49.6                                 \\
                             & AutoTrack~\cite{li2020autotrack}     & CVPR 20                  & 71.8                                 & 45.0                                 & 78.8                                 & 57.3                                 & 68.9                                 & 47.2                                 & 67.1                                 & 47.7                                 & 71.7                                 & 49.3                                 \\
\multirow{-8}{*}{\rotatebox{90}{DCF-based}}  & RACF~\cite{li2022learning}          & PR 20                    & 77.3                                 & 49.4                                 & 83.4                                 & 60.0                                 & 70.2                                 & 47.7                                 & 69.4                                 & 48.6                                 & 75.1                                 & 51.4                                 \\ \hline
                             & HiFT~\cite{cao2021hift}          & ICCV 21                  & 65.2                                 & 47.5                                 & 71.9                                 & 52.6                                 & 78.7                                 & 59.0                                 & 74.9                                 & 57.0                                 & 72.7                                 & 54.0                                 \\
                             & SiamAPN++~\cite{cao2021siamapn++}     & IROS 21                  & 76.9                                 & 55.6                                 & 73.5                                 & 53.2                                 & 76.8                                 & 58.2                                 & 76.4                                 & 58.1                                 & 75.9                                 & 56.3                                 \\
                             & LightTrack~\cite{Yan2021LightTrackFL}    & CVPR 21                  & 80.4                                 & 61.2                                 & 74.8                                 & 56.8                                 & 78.3                                 & 62.7                                 & 75.1                                 & 59.9                                 & 77.2                                 & 60.2                                 \\
                             & TCTrack~\cite{cao2022tctrack}       & CVPR 22                  & 69.1                                 & 50.4                                 & 77.6                                 & 57.7                                 & 77.3                                 & 60.4                                 & 75.1                                 & 58.8                                 & 74.7                                 & 56.8                                 \\
                             & TCTrack++~\cite{cao2023towards}     & TPAMI 23                 & 72.5                                 & 53.2                                 & 80.8                                 & 60.3                                 & 74.4                                 & 58.8                                 & 78.2                                 & 60.1                                 & 76.5                                 & 58.1                                 \\
                             & SGDViT~\cite{yao2023sgdvit}        & ICRA 23                  & 65.7                                 & 48.0                                 & 72.1                                 & 52.1                                 & 75.4                                 & 57.5                                 & {\color[HTML]{FE0000} \textBF{86.3}} & {\color[HTML]{FE0000} \textBF{66.1}} & 74.9                                 & 55.9                                 \\
                             & ABDNet~\cite{zuo2023adversarial}        & RAL 23                   & 75.5                                 & 55.3                                 & 75.0                                 & 57.2                                 & 79.3                                 & 60.7                                 & 77.3                                 & 59.1                                 & 76.8                                 & 58.1                                 \\
  & DRCI~\cite{zeng2023towards}          & ICME 23                  & {\color[HTML]{009901} \textBF{83.0}} & {\color[HTML]{009901} \textBF{59.0}} & 83.4                                 & 60.0                                 & 76.7                                 & 59.7                                 & 73.6                                 & 55.2                                 & 79.2                                 & 58.5                                 \\
\multirow{-9}{*}{\rotatebox{90}{CNN-based}} & PRL-Track~\cite{fu2024progressive}                                      & IROS 24                                                 & 73.1                                                         & 53.5                                                         & 72.6                                                         & 53.8                                                         & 79.1                                                         & 59.3                                                         & 74.1                                                         & 58.6                                                         & 74.7                                                         & 56.3                                                         \\ \hline
                             & Aba-ViTrack~\cite{li2023adaptive}     & ICCV 23                  & {\color[HTML]{3531FF} \textBF{83.4}} & {\color[HTML]{3531FF} \textBF{59.9}} & {\color[HTML]{3531FF} \textBF{86.1}} & {\color[HTML]{FE0000} \textBF{65.3}} & {\color[HTML]{FE0000} \textBF{86.4}} & {\color[HTML]{009901} \textBF{66.4}} & {\color[HTML]{3531FF} \textBF{85.0}} & {\color[HTML]{3531FF} \textBF{65.5}} & {\color[HTML]{FE0000} \textBF{85.2}} & {\color[HTML]{3531FF} \textBF{64.3}} \\
                             & LiteTrack~\cite{Wei2023LiteTrackLP}     & ICRA 24                  & 81.6                                 & 59.3                                 & 79.7                                 & 61.4                                 & {\color[HTML]{009901} \textBF{84.2}} & 65.9                                 & 83.1                                 & 65.0                                 & 82.2                                 & 62.9                                 \\
                             & SMAT~\cite{gopal2024separable}          & WACV 24                  & 80.8                                 & 58.7                                 & 82.5                                 & 63.4                                 & 81.8                                 & 64.6                                 & 80.4                                 & 63.5                                 & 81.4                                 & 62.6                                 \\
                             & LightFC~\cite{Li2024LightweightFS}       & KBS 24                   & 83.4                                 & 60.6                                 & 82.7                                 & 62.8                                 & {\color[HTML]{009901} \textBF{84.2}} & 65.5                                 & 81.3                                 & 63.7                                 & 82.9                                 & 63.1                                 \\
                             & AVTrack-DeiT~\cite{lilearning2024}  & ICML 24                   & 82.1                                 & 58.7                                 & {\color[HTML]{FE0000} \textBF{86.0}} & {\color[HTML]{FE0000} \textBF{65.3}} & {\color[HTML]{3531FF} \textBF{84.8}} & {\color[HTML]{FE0000} \textBF{66.8}} & 83.2                                 & {\color[HTML]{009901} \textBF{65.8}} & {\color[HTML]{009901} \textBF{84.0}} & {\color[HTML]{009901} \textBF{64.1}} \\
                             & \cellcolor[HTML]{DEF3FE}\textBF{BDTrack-ViT}   &  \cellcolor[HTML]{DEF3FE}                    & \cellcolor[HTML]{DEF3FE}78.9                                 & \cellcolor[HTML]{DEF3FE}57.3                                 & \cellcolor[HTML]{DEF3FE}83.9                                 & \cellcolor[HTML]{DEF3FE}63.6                                 & \cellcolor[HTML]{DEF3FE}83.5                                 & \cellcolor[HTML]{DEF3FE}65.5                                 & \cellcolor[HTML]{DEF3FE}{\color[HTML]{009901} \textBF{84.1}} & \cellcolor[HTML]{DEF3FE}{\color[HTML]{FE0000} \textBF{66.1}} & \cellcolor[HTML]{DEF3FE}82.6                                 & \cellcolor[HTML]{DEF3FE}63.1                                 \\
\multirow{-7}{*}{\rotatebox{90}{ViT-based}}  & \cellcolor[HTML]{DEF3FE}\textBF{BDTrack-DeiT}  & \cellcolor[HTML]{DEF3FE}\multirow{-2}{*}{\textBF{Ours}}                         & \cellcolor[HTML]{DEF3FE}{\color[HTML]{FE0000} \textBF{84.1}} & \cellcolor[HTML]{DEF3FE}{\color[HTML]{FE0000} \textBF{61.0}} & \cellcolor[HTML]{DEF3FE}{\color[HTML]{009901} \textBF{85.2}} & \cellcolor[HTML]{DEF3FE}{\color[HTML]{009901} \textBF{64.3}} & \cellcolor[HTML]{DEF3FE}{\color[HTML]{3531FF} \textBF{84.8}} & \cellcolor[HTML]{DEF3FE}{\color[HTML]{3531FF} \textBF{66.7}} & \cellcolor[HTML]{DEF3FE}83.5                                 & \cellcolor[HTML]{DEF3FE}65.9                                 & \cellcolor[HTML]{DEF3FE}{\color[HTML]{3531FF} \textBF{84.4}} & \cellcolor[HTML]{DEF3FE}{\color[HTML]{FE0000} \textBF{64.5}} \\ \bottomrule[1pt]
\end{tabular}
\end{table}

\section{Experiments}\label{section_experiment}
This section provides detailed and comprehensive evaluation results of our approach across four well-known UAV tracking benchmarks: namely, UAV123 \cite{mueller2016benchmark}, UAV123@10fps \cite{mueller2016benchmark}, VisDrone2018 \cite{wen2018visdrone}, and UAVDT \cite{du2018unmanned}.
A PC equipped with a NVIDIA TitanX GPU, 16GB RAM, and an i9-10850K CPU (3.6GHz) was used for all evaluation experiments.
In order to provide a comprehensive evaluation,  we evaluate our method against 22 lightweight SOTA trackers,  which are categorized into three groups as in Aba-ViTrack~\cite{li2023adaptive}, i.e., DCF-based, CNN-based, and ViT-based methods (see Table  \ref{table_lightweight}), alongside 14 SOTA deep trackers tailored for generic visual tracking (refer to Table \ref{table-deep-trackers}). 
Note that in our study, ``light trackers" and ``efficient trackers" are used interchangeably to refer to trackers that prioritize low computational complexity and reduced memory usage, typically designed for environments with limited resources or real-time applications.

\subsection{Implementation Details}

We employed three efficient ViT models, i.e., ViT-tiny \cite{dosovitskiy2020image} and DeiT-tiny \cite{touvron2021training}, as the backbones to develop three distinct trackers, named BDTrack-ViT, BDTrack-DeiT, and BDTrack-T2T, respectively.
The sizes of the search region and template are set to $256^2$ and $128^2$.
The head architecture, training data, hyperparameters, and training pipeline adhere to the specifications outlined in Aba-ViTrack \cite{li2023adaptive}. And we apply Hanning window penalties during inference to integrate positional priors in tracking, following established practices \cite{zhang2020ocean}.

\subsection{Comparison with Lightweight Trackers}
In this section, we comprehensively evaluate BDTrack’s performance by comparing it with 22 SOTA lightweight trackers across four UAV tracking benchmarks. The evaluation results are shown in Table \ref{table_lightweight}.
As can be seen, our BDTrack-DeiT outperforms other trackers except Aba-ViTrack on all benchmarks, regarding average (Avg.) precision (Prec.) and success rate (Succ.).
RACF \cite{li2022learning} secures the highest  Prec. and Succ. in average among DCF-based trackers, with values of 75.1\% and 51.4\%, respectively. Among CNN-based trackers, DRCI \cite{zeng2023towards} stands out with the highest average Prec. of 79.2\% and Succ. of 58.5\%,. 
However, all ViT-based trackers exhibit Avg. Prec. and Succ.  exceeding 81.0\% and 62.0\%, respectively, obviously beating the CNN-based methods and substantially outperforming the DCF-based ones. 
Although Aba-ViTrack achieves the highest Avg. Prec. (85.2\%), our BDTrack-DeiT ranks second with only a slight gap of 0.8\%, while maintaining the best Avg. Succ. of 64.5\%.
Notably, BDTrack-DeiT outperforms Aba-ViTrack on the UAVDT, with a success rate improvement of over 1.0\%.
These results highlight the advantages of our method and justify its SOTA performance for UAV tracking.

\subsection{Comparison with Deep Trackers}
The proposed BDTrack-DeiT is also compared with 14 SOTA deep trackers.
Table \ref{table-deep-trackers} presents the results in terms of precision (Prec.), success rate (Succ.), and their average (Avg.).
It is worth noting that most deep trackers are tailored for generic tracking scenarios without prioritizing real-time constraints. Consequently, they tend to exhibit slower tracking speeds and higher model complexity, as shown in Table \ref{table_paramaters}.
Our BDTrack-DeiT stands out by offering the fastest speed without compromising on high performance, showcasing its competitive edge in terms of both performance and speed. 
Remarkably, our method achieves third place in Avg. precision on four benchmarks while securing the second position in precision on UAVDT and VisDrone2018. This underscores the effectiveness of our approach for UAV tracking tasks and highlights that our BDTrack-DeiT is equivalent in precision to state-of-the-art deep trackers.
Although other deep trackers like ROMTrack \cite{cai2023robust} and EVPTrack \cite{shi2024evptrack} may achieve comparable performance with ours, their GPU speeds are significantly slower. In fact, our method outperforms SeqTrack and EVPTrack by being 5 and 11 times faster, respectively.
By showcasing its capability to offer high precision and speed, these results demonstrate the suitability of our approach for real-time UAV tracking applications that prioritize accuracy and efficiency.

\begin{table}[t]
\centering
\scriptsize
\setlength\tabcolsep{2.5pt} 
\caption{Precision (Prec.) and success rate (Succ.) comparison between BDTrack-DeiT and DL-based trackers.}
\label{table-deep-trackers}
\begin{tabular}{cccccccccccc}
\toprule[1pt]
                                              &                                       & \multicolumn{2}{c}{UAVDT}                                                   & \multicolumn{2}{c}{VisDrone2018}                                            & \multicolumn{2}{c}{UAV123}                                                  & \multicolumn{2}{c}{UAV123@10fps}                                            & \multicolumn{2}{c}{Avg.}                                                    \\
\multirow{-2}{*}{Tracker}                     & \multirow{-2}{*}{Source}              & Prec.                                & Succ.                                & Prec.                                & Succ.                                & Prec.                                & Succ.                                & Prec.                                & Succ.                                & Prec.                                & Succ.                                \\ \hline
PrDiMP~\cite{danelljan2020probabilistic}                                        & CVPR 20                               & 75.8                                 & 55.9                                 & 79.8                                 & 60.2                                 & 87.2                                 & 66.5                                 & 83.9                                 & 64.7                                 & 81.7                                 & 61.8                                 \\
SOAT~\cite{zhou2021saliency}                                          & CVPR 21                               & 82.1                                 & 60.7                                 & 76.9                                 & 59.7                                 & 82.7                                 & 64.9                                 & 85.2                                 & 65.7                                 & 81.7                                 & 62.8                                 \\
AutoMatch~\cite{zhang2021learn}                                     & ICCV 21                               & 82.1                                 & 62.9                                 & 78.1                                 & 59.6                                 & 83.8                                 & 64.4                                 & 78.1                                 & 59.4                                 & 80.5                                 & 61.6                                 \\
SparseTT~\cite{fu2022sparsett}                                      & IJCAI 22                              & {\color[HTML]{009901} \textBF{82.8}} & {\color[HTML]{FE0000} \textBF{65.4}} & 81.4                                 & 62.1                                 & 85.4                                 & 68.8                                 & 82.2                                 & 64.9                                 & 82.9                                 & 65.3                                 \\
CSWinTT~\cite{song2022transformer}                                       & CVPR 22                               & 67.3                                 & 54.0                                 & 75.2                                 & 58.0                                 & 87.6                                 & {\color[HTML]{FE0000} \textBF{70.5}} & 87.1                                 & 68.1                                 & 79.3                                 & 62.7                                 \\
SimTrack~\cite{chen2022backbone}                                      & ECCV 22                               & 76.5                                 & 57.2                                 & 80.0                                 & 60.9                                 & {\color[HTML]{009901} \textBF{88.2}} & 69.2                                 & {\color[HTML]{009901} \textBF{87.5}} & 69.0                                 & 83.1                                 & 64.0                                 \\
OSTrack~\cite{ye2022joint}                                       & ECCV 22                               & {\color[HTML]{FE0000} \textBF{85.1}} & {\color[HTML]{3531FF} \textBF{63.4}} & 84.2                                 & {\color[HTML]{009901} \textBF{64.8}} & 84.7                                 & 67.4                                 & 83.1                                 & 66.1                                 & 84.2                                 & 65.2                                 \\
ZoomTrack~\cite{kou2023zoomtrack}                                     & NIPS 23                               & 77.1                                 & 57.9                                 & 81.4                                 & 63.6                                 & {\color[HTML]{3531FF} \textBF{88.4}} & {\color[HTML]{009901} \textBF{69.6}} & {\color[HTML]{FE0000} \textBF{88.8}} & {\color[HTML]{3531FF} \textBF{70.0}} & 83.9                                 & {\color[HTML]{009901} \textBF{65.3}} \\
SeqTrack~\cite{chen2023seqtrack}                                      & CVPR 23                               & 78.7                                 & 58.8                                 & 83.3                                 & 64.1                                 & 86.8                                 & 68.6                                 & 85.7                                 & 68.1                                 & 83.6                                 & 64.9                                 \\
MAT~\cite{zhao2023representation}                                           & CVPR 23                               & 72.9                                 & 54.8                                 & 81.6                                 & 62.2                                 & 86.7                                 & 68.3                                 & 86.9                                 & 68.5                                 & 82.0                                 & 63.5                                 \\
ROMTrack~\cite{cai2023robust}                                      & ICCV 23                               & 81.9                                 & {\color[HTML]{009901} \textBF{61.6}} & {\color[HTML]{FE0000} \textBF{86.4}} & {\color[HTML]{FE0000} \textBF{66.7}} & 87.4                                 & 69.2                                 & 85.0                                 & {\color[HTML]{333333} 67.8}          & {\color[HTML]{3531FF} \textBF{85.2}} & {\color[HTML]{3531FF} \textBF{66.3}} \\
DCPT~\cite{zhu2024dcpt}                                          & ICRA 24                               & 76.8                                 & 56.9                                 & 83.1                                 & 64.2                                 & 85.7                                 & 68.1                                 & 86.9                                 & {\color[HTML]{009901} \textBF{69.1}} & 83.1                                 & 64.5                                 \\
EVPTrack~\cite{shi2024evptrack}                                      & AAAI 24                               & 80.6                                 & 61.2                                 & {\color[HTML]{009901} \textBF{84.5}} & {\color[HTML]{3531FF} \textBF{65.8}} & {\color[HTML]{FE0000} \textBF{88.9}} & {\color[HTML]{3531FF} \textBF{70.2}} & {\color[HTML]{3531FF} \textBF{88.7}} & {\color[HTML]{FE0000} \textBF{70.4}} & {\color[HTML]{FE0000} \textBF{85.7}} & {\color[HTML]{FE0000} \textBF{66.9}} \\
SuperSBT~\cite{xie2024correlation}                                      & TPAMI 24                              & 80.1                                 & 59.2                                 & 82.2                                 & 63.5                                 & 85.3                                 & 67.3                                 & 87.3                                 & {\color[HTML]{009901} \textBF{69.1}} & 83.7                                 & 64.7                                 \\
\rowcolor[HTML]{DEF3FE} 
\cellcolor[HTML]{DEF3FE}\textBF{BDTrack-DeiT} & \cellcolor[HTML]{DEF3FE}\textBF{Ours} & {\color[HTML]{3531FF} \textBF{84.1}} & \cellcolor[HTML]{DEF3FE}61.0         & {\color[HTML]{3531FF} \textBF{85.2}} & {\color[HTML]{333333} 64.3}          & {\color[HTML]{333333} 84.8}          & {\color[HTML]{333333} 66.7}          & 83.5                                 & 65.9                                 & {\color[HTML]{009901} \textBF{84.4}} & {\color[HTML]{333333} 64.5}   \\ \bottomrule[1pt]      
\end{tabular}
\end{table}

\begin{table}[h]
\centering
\scriptsize
\setlength\tabcolsep{3.5pt}
\caption{Comparison of FLOPS, parameters (Params.), and average speed between lightweight and deep trackers on four UAV tracking benchmarks.}
\label{table_paramaters}
\begin{tabular}{cccccc}
\toprule[1pt]
\multicolumn{2}{c}{}                                                                &                                                                       &                                                                         & \multicolumn{2}{c}{Avg.FPS}                                                                  \\
\multicolumn{2}{c}{\multirow{-2}{*}{Method}}                                        & \multirow{-2}{*}{\begin{tabular}[c]{@{}c@{}}FLOPs\\ (G)\end{tabular}} & \multirow{-2}{*}{\begin{tabular}[c]{@{}c@{}}Params.\\ (M)\end{tabular}} & GPU                                                           & CPU                          \\ \hline
                              & HiFT~\cite{cao2021hift}                                                & 7.2                                                                   & 9.9                                                                     & 157.6                                                         & -                            \\
                              & SiamAPN++~\cite{cao2021siamapn++}                                           & 8.2                                                                   & 14.7                                                                    & 162.3                                                         & -                            \\
                              & TCTrack~\cite{cao2022tctrack}                                             & 16.9                                                                  & 8.5                                                                     & 133.8                                                         & -                            \\
                              & SGDViT~\cite{yao2023sgdvit}                                              & 11.3                                                                  & 23.3                                                                    & 105.9                                                         & -                            \\
                              & PRL-Track~\cite{fu2024progressive}                                            & 7.4                                                                   & 12.1                                                                    & 133.2                                                         & -                            \\
                              & Aba-ViTrack~\cite{li2023adaptive}                                         & 2.4                                                                   & 7.9                                                                     & {\color[HTML]{009901} \textBF{177.8}}                         & 49.7                         \\
                              & LiteTrack~\cite{Wei2023LiteTrackLP}                                           & 28.3                                                                  & 7.3                                                                     & 115.6                                                         & -                            \\
                              & AVTrack-DeiT~\cite{lilearning2024}                                        & 0.97-2.4                                                              & 3.5-7.9                                                                 & {\color[HTML]{3531FF} \textBF{253.6}}                         & 58.8                         \\
\multirow{-9}{*}{\rotatebox{90}{Lightweight}} & \cellcolor[HTML]{DEF3FE}\textBF{BDTrack-DeiT(Ours)} & \cellcolor[HTML]{DEF3FE}{\color[HTML]{333333} 1.7-2.4}                & \cellcolor[HTML]{DEF3FE}{\color[HTML]{333333} 5.8-7.9}                  & \cellcolor[HTML]{DEF3FE}{\color[HTML]{FE0000} \textBF{283.4}} & \cellcolor[HTML]{DEF3FE}63.2 \\ \hline
                              & OSTrack~\cite{ye2022joint}                                             & 21.5                                                                  & 92.1                                                                    & 65.4                                                          & -                            \\
                              & SeqTrack~\cite{chen2023seqtrack}                                            & 66.3                                                                  & 89.4                                                                    & 30.1                                                          & -                            \\
                              & ZoomTrack~\cite{kou2023zoomtrack}                                           & 21.2                                                                  & 91.8                                                                    & 60.8                                                          & -                            \\
                              & MAT~\cite{zhao2023representation}                                                 & 42.9                                                                  & 88.4                                                                    & 68.4                                                          & -                            \\
                              & ROMTrack~\cite{cai2023robust}                                            & 34.5                                                                  & 92.1                                                                    & 50.8                                                          & -                            \\
                              & DCPT~\cite{zhu2024dcpt}                                                & 29.4                                                                  & 92.9                                                                    & 37.3                                                          & -                            \\
                              & EVPTrack~\cite{shi2024evptrack}                                            & 65.4                                                                  & 73.5                                                                    & 24.1                                                          &  -                            \\
\multirow{-8}{*}{\rotatebox{90}{Deep}}        & SuperSBT~\cite{xie2024correlation}                                            & 24.6                                  & 65.5                                   & 30.3                                                          &-     \\ \bottomrule[1pt]                        
\end{tabular}
\end{table}

\subsection{Efficiency Comparison}

To highlight the efficiency of the proposed trackers compared to existing methods, Table \ref{table_paramaters} presents the floating-point operations (FLOPs), the number of parameters during inference (Param.), and the inference speed of our method, alongside 8 SOTA efficient lightweight trackers and 8 SOTA deep trackers.
Notably, as our method and AVTrack-DeiT feature adaptive architectures, the FLOPs and Params. of them vary within a range, spanning from the minimum to the maximum values.
As observed, our methods achieve significantly lower FLOPs and Params. compared to all rival trackers. Even our maximum values are lower than those of most lightweight trackers. 
In terms of GPU speed, BDTrack-DeiT achieves the highest performance at 283.4 FPS, followed closely by AVTrack-DeiT~\cite{lilearning2024} with 253.6 FPS.
As for CPU speed, our tracker exhibits real-time performance on a single CPU\footnote{Note that the real-time performance discussed in this paper can be generalized only to platforms similar to or more advanced than ours.}.
Although Aba-ViTrack~\cite{li2023adaptive} achieves comparable performance to BDTrack-DeiT, it exhibits significantly slower inference speed.
Remarkably, BDTrack-DeiT boasts over 1.6 times the GPU speed and 1.3 times the CPU speed of Aba-ViTrack, striking a better balance between tracking precision and efficiency. 
On the other hand, although existing deep trackers achieve comparable performance to ours, they suffer from significantly slower speeds and higher model complexity. In contrast, our method outperforms ROMTrack~\cite{cai2023robust} and EVPTrack~\cite{shi2024evptrack} by factors of 5 and 11, respectively. These results further highlight the suitability of our approach for real-time UAV tracking applications where both accuracy and efficiency are critical.

\begin{figure}[t]
    \centering
    \includegraphics[width=0.47\textwidth,height=0.25\textwidth]{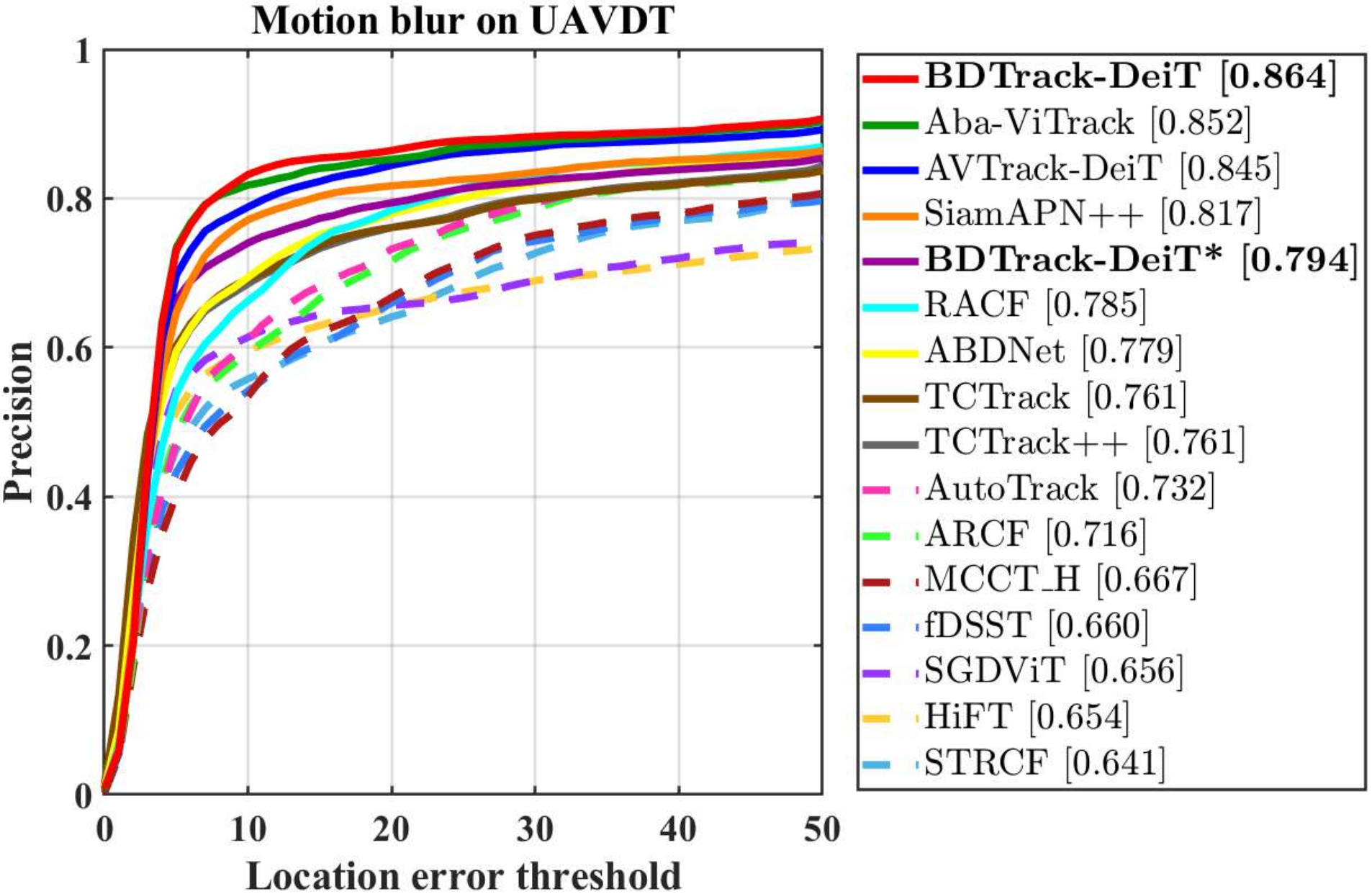}
    \includegraphics[width=0.47\textwidth,height=0.25\textwidth]{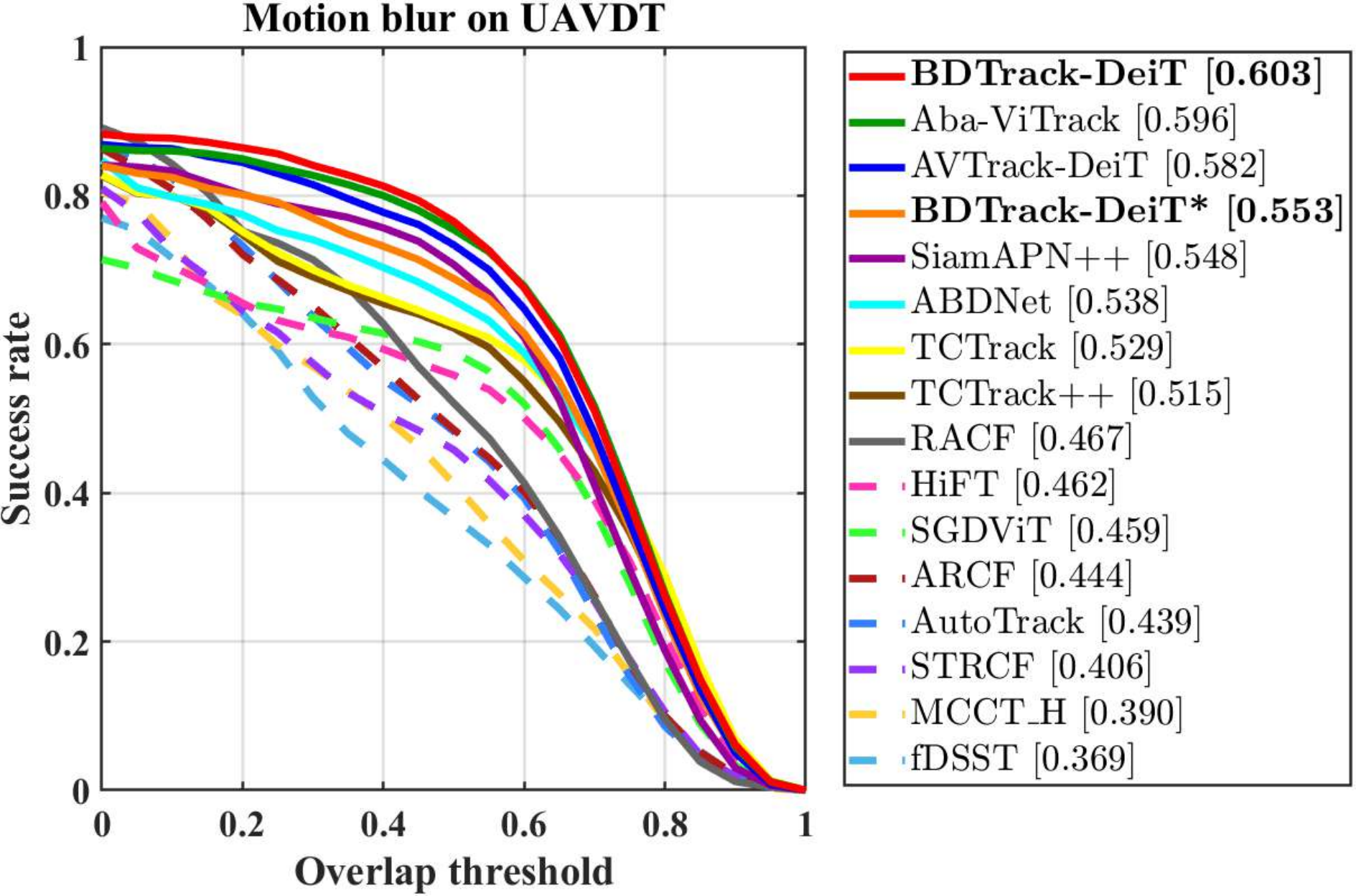}

    \vskip 1ex 

    \includegraphics[width=0.47\textwidth,height=0.25\textwidth]{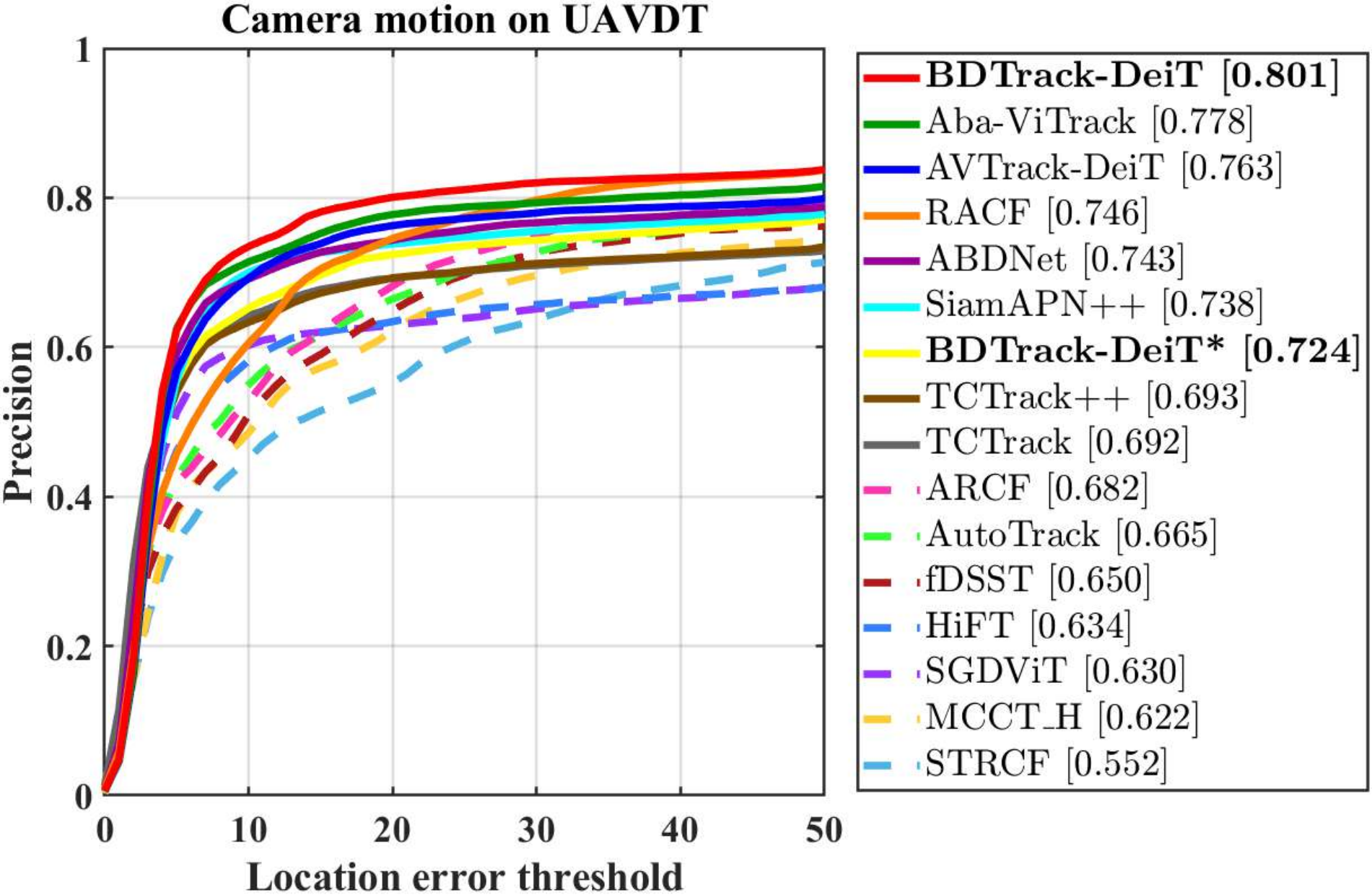}
    \includegraphics[width=0.47\textwidth,height=0.25\textwidth]{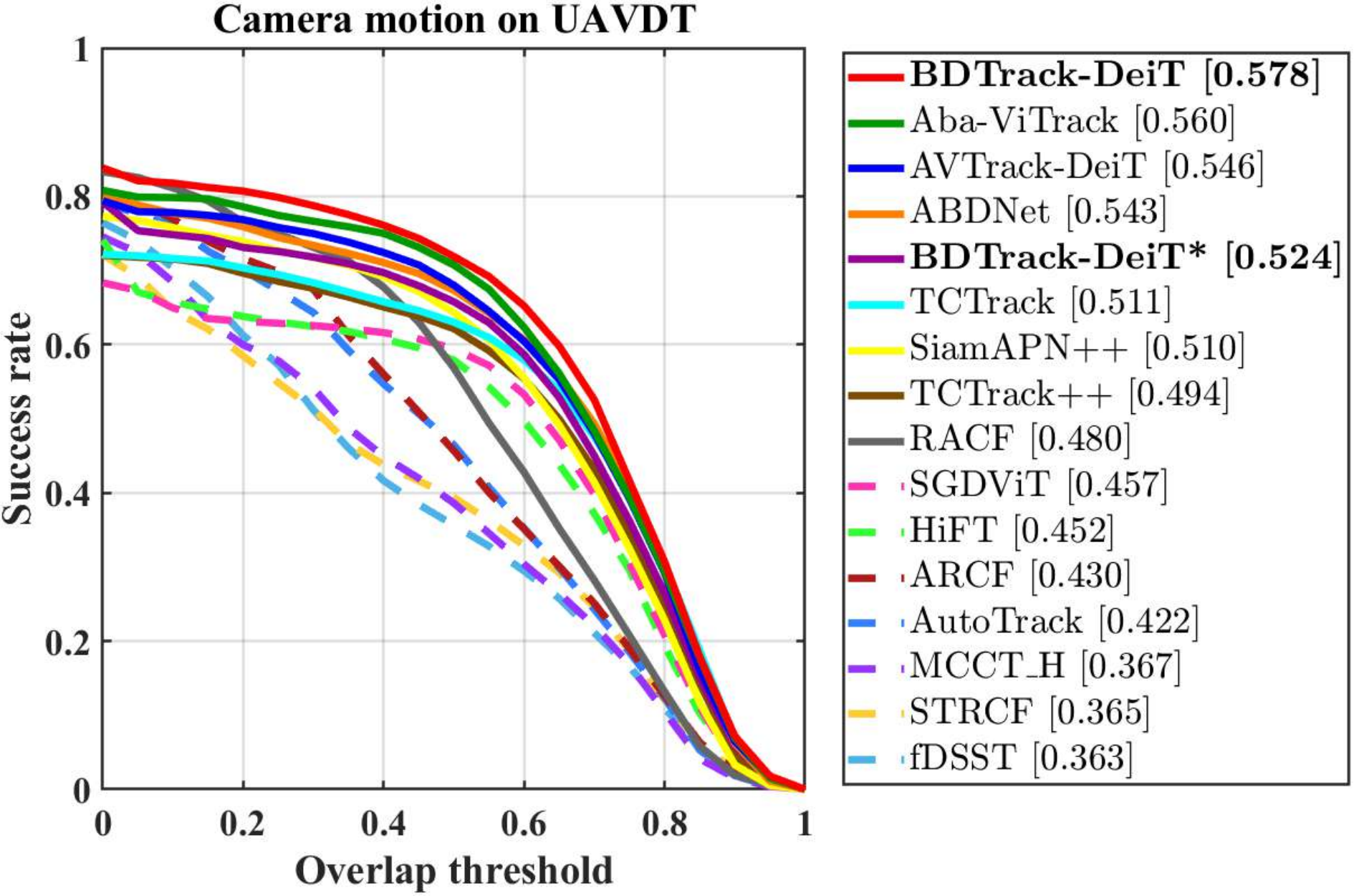}
\caption{The precision and success plots for attribute-based comparisons on `Motion Blur' (top row) and `Camera Motion' (bottom row) are presented for the attribute subsets of UAVDT.
Note that \textbf{BDTrack-DeiT*} denotes \textbf{BDTrack-DeiT} without the application of the proposed components.}
\label{fig_attr_plots}
\end{figure}

    


 \subsection{Attribute-Based Evaluation}
To further validate the superiority of the proposed tracker over existing UAV trackers, we evaluate the performance of BDTrack-DeiT by comparing it with 14 SOTA UAV trackers on three subsets of the UAVDT~\cite{du2018unmanned}, including `Motion blur' and `Camera motion'.
Note that we also evaluate BDTrack-DeiT without applying the components (i.e., MBRV and DEEM), denoted as BDTrack-DeiT* for reference.
The precision and success plots are illustrated in Fig. \ref{fig_attr_plots}. 
As observed, BDTrack-DeiT significantly outperforms the second tracker on `Motion blur' and `Camera motion', with gains of 1.2\%/0.7\% and 2.3\%/1.8\% in precision/success rate, respectively.
Significantly, across all these two attributes, the incorporation of the proposed components yields a noteworthy enhancement over BDTrack-DeiT*, with improvements of 7.0\%/5.0\% and 7.9\%/5.5\%in terms of precision and success rate.
Note that `Camera motion' is the factor that can contribute to motion blur in visual tracking systems. When a camera moves rapidly or changes direction suddenly, it can cause motion blur in the captured images. 
Therefore, this attribute-based evaluation validates, directly or indirectly, the superiority and effectiveness of the proposed method in addressing the challenges posed by motion blur.

\begin{figure*}[t]
\centering
\includegraphics[width=0.95\textwidth]{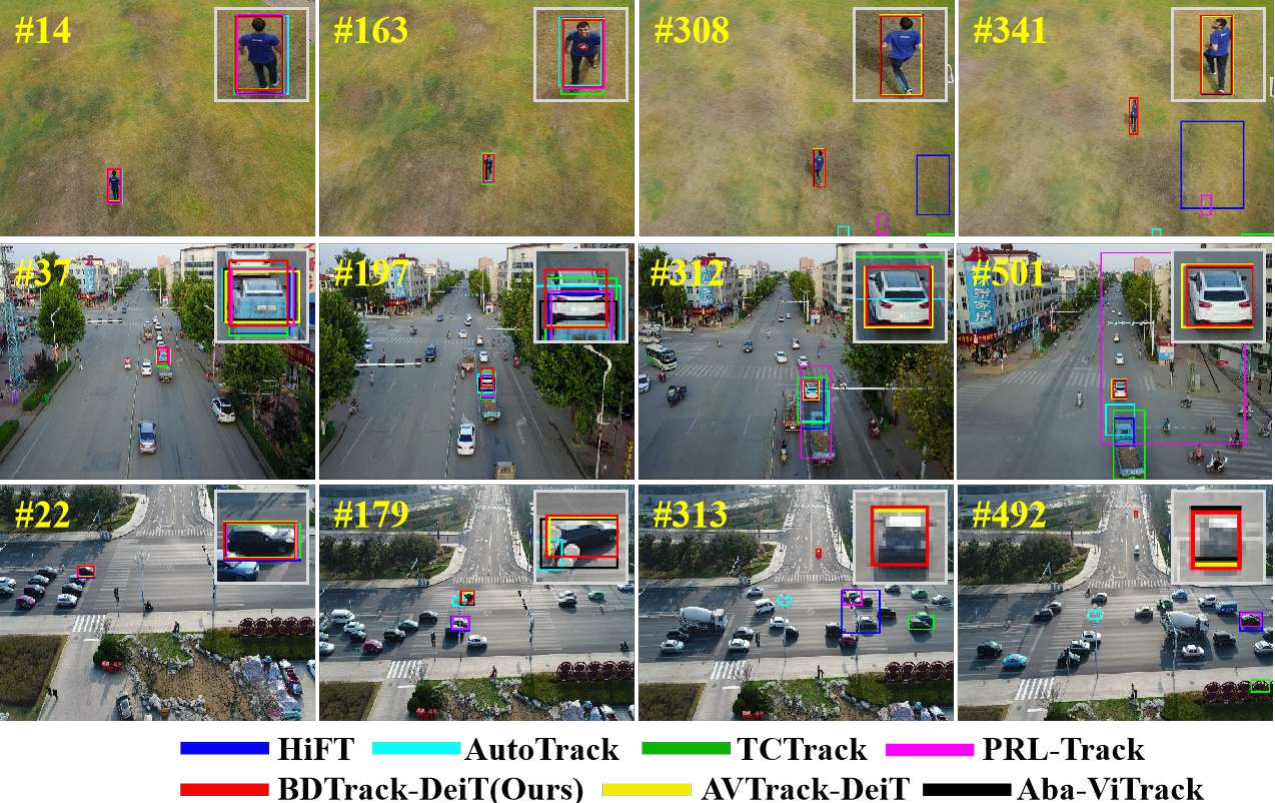}
\caption{Visualization of qualitative comparisons between our tracker and six SOTA UAV trackers on three sequences from UAV123~\cite{mueller2016benchmark}, VisDrone2018~\cite{wen2018visdrone}, and UAVDT~\cite{du2018unmanned} (i.e., person10, uav0000164\_00000\_s, and S1606).}
 \label{BDTrack-visual-examples}
\end{figure*}

\subsection{Qualitative evaluation}

Qualitative tracking results of BDTrack-DeiT and six SOTA trackers, i.e., HiFT \cite{cao2021hift}, AutoTrack \cite{li2020autotrack}, TCTrack \cite{cao2022tctrack}, PRL-Track \cite{fu2024progressive}, AVTrack-DeiT \cite{lilearning2024}, and Aba-ViTrack \cite{li2023adaptive}, are presented in Fig. \ref{BDTrack-visual-examples}. 
Three video sequences from three different benchmarks are selected for demonstration: person10 from UAV123, uav0000164\_00000\_s from VisDrone2018, and S1606 from UAVDT.
For better visualization, the low-resolution targets within the video frames are selectively zoomed, cropped, and placed in the upper right corner of their corresponding frames for display.
As observed, our tracker can successfully tracks the targets in all challenging examples, featuring camera motion (i.e, in all sequences), occlusion (i.e, uav0000164\_00000\_s), and background clutter (i.e, S1606). 
Our method not only exhibits superior performance but also delivers more visually pleasing results in these scenarios, further validating the effectiveness of the proposed method for UAV tracking.

\begin{table}[h]
\scriptsize
\setlength\tabcolsep{2.5pt} 
\centering
\caption{Ablation study on the impact of MBRV and DEEM on the baseline tracker's performance.}
\label{table_backbone_ablation}
\begin{tabular}{cccccccccccc}
\toprule[1pt]
\multirow{2}{*}{Tracker}               & \multirow{2}{*}{MBVR} & \multirow{2}{*}{DEEM} & \multicolumn{2}{c}{UAVDT}     & \multicolumn{2}{c}{VisDrone2018}  & \multicolumn{2}{c}{UAV123}    & \multicolumn{2}{c}{Avg}       & \multirow{2}{*}{Avg.FPS} \\
                                       &                       &                       & Prec.         & Succ.         & Prec.         & Succ.         & Prec.         & Succ.         & Prec.         & Succ.         &                          \\ \hline
\multirow{3}{*}{\textBF{BDTrack-DeiT}} &                       &                       & 78.6          & 56.7          & 81.6          & 62.2          & 83.7          & 66.1          & 81.3          & 61.7          & 228.7                    \\
                                       &$\checkmark$                       &                      & \textBF{84.5} & \textBF{61.4} & \textBF{86.1} & \textBF{65.1} & \textBF{85.3} & \textBF{67.0} & \textBF{85.3} & \textBF{64.5} & -                        \\
                                       &$\checkmark$                       &$\checkmark$                       & 84.1          & 61.0          & 85.2          & 64.3          & 84.8          & 66.7          & 84.7          & 64.0          & \textBF{279.4}$_{\uparrow 22.3\%}$      \\ \bottomrule[1pt]    
\end{tabular}
\end{table}

\subsection{Ablation Study}
\subsubsection{Impact of learning blur robust ViTs with dynamic early exit}

We add the proposed components MBRV and DEEM into the baselines one by one to evaluate their effectiveness. The evaluation results on three benchmarks are shown in Table \ref{table_backbone_ablation}.
As GPU speed of the baseline and its enhanced version by MBRV are theoretically equal, we provide only the speed of the baseline to eliminate potential nuances arising from randomness.
As can be seen, the inclusion of MBRV leads to significant improvements in both Prec. and Succ. for the baseline tracker. 
Specifically, the inclusion of the MBRV leads to a significant increase of over 4.0\% in average Prec. and 2.8\% in average Succ. for the baseline.
The improvement in performance after adding the MSE loss can be attributed to the enhanced feature consistency between clear and blurred templates. By enforcing feature similarity, the model learns more robust representations that generalize well to different blur conditions and focuses on discriminative features, leading to overall better performance beyond just handling blurred scenes.
When DEEM is further integrated, consistent improvements in GPU speeds are observed, accompanied by only slight decreases in Prec. and Succ.
Specifically, the GPU speeds of all baselines see increases of above 22.0\%. 
Despite the marginal drop in Prec. and Succ. (the average drops are less than 0.6\%), the overall pattern of enhanced GPU speeds and minor sacrifices in tracking performance validates the effectiveness of DEEM in optimizing tracking efficiency.

\begin{table}[]
\centering
\scriptsize
\setlength\tabcolsep{2.5pt}
\caption{Ablation study on the weighting of loss $\mathcal{L}_{br}$ in training BDTrack-DeiT  by varying $\rho$ from $0.5 \times 10^{-4}$ to $5.0 \times 10^{-4}$.}
\label{table-loss-blur}
\begin{tabular}{ccccccccccc}
\toprule[1pt]
                       & \multicolumn{2}{c}{UAVDT}                                                   & \multicolumn{2}{c}{VisDrone2018}                                            & \multicolumn{2}{c}{UAV123}                                                  & \multicolumn{2}{c}{UAV123@10fps}                                            & \multicolumn{2}{c}{Avg.}                                                     \\
\multirow{-2}{*}{$\rho \times 10^{-4}$} & Prec.                                & Succ.                                & Prec.                                & Succ.                                & Prec.                                & Succ.                                & Prec.                                & Succ.                                & Prec.                                & Succ.                                \\ \hline
0.5                    & {\color[HTML]{009901} \textBF{83.4}} & {\color[HTML]{009901} \textBF{60.0}} & 83.7                                 & 63.5                                 & 83.1                                 & 65.4                                 & 83.2                                 & 65.6                                 & 83.3                                 & 63.6                                 \\
0.6                    & 82.3                                 & 59.3                                 & 84.5                                 & 63.9                                 & 83.7                                 & 66.0                                 & 83.0                                 & 65.4                                 & 83.4                                 & 63.6                                 \\
0.7                    & 83.0                                 & 59.6                                 & 85.8                                 & 64.6                                 & {\color[HTML]{009901} \textBF{84.1}} & {\color[HTML]{009901} \textBF{66.5}} & {\color[HTML]{009901} \textBF{83.3}} & {\color[HTML]{009901} \textBF{66.2}} & {\color[HTML]{3531FF} \textBF{84.0}} & {\color[HTML]{3531FF} \textBF{64.2}} \\
0.8                    & 83.2                                 & 59.8                                 & 81.1                                 & 61.9                                 & 83.5                                 & {\color[HTML]{009901} \textBF{65.8}} & 82.3                                 & 64.9                                 & 82.5                                 & 63.1                                 \\
0.9                    & 82.5                                 & 59.7                                 & 84.6                                 & 63.9                                 & 84.0                                 & 66.1                                 & 83.1                                 & 65.6                                 & 83.6                                 & 63.8                                 \\
1.0                    & {\color[HTML]{FE0000} \textBF{84.1}} & {\color[HTML]{FE0000} \textBF{61.0}} & {\color[HTML]{009901} \textBF{85.2}} & {\color[HTML]{009901} \textBF{64.3}} & {\color[HTML]{FE0000} \textBF{84.8}} & {\color[HTML]{FE0000} \textBF{66.7}} & {\color[HTML]{3531FF} \textBF{83.5}} & {\color[HTML]{3531FF} \textBF{65.9}} & {\color[HTML]{FE0000} \textBF{84.4}} & {\color[HTML]{FE0000} \textBF{64.5}} \\
2.0                    & 80.4                                 & 58.2                                 & {\color[HTML]{FE0000} \textBF{86.1}} & {\color[HTML]{FE0000} \textBF{64.8}} & 84.0                                 & 65.9                                 & {\color[HTML]{FE0000} \textBF{84.9}} & {\color[HTML]{FE0000} \textBF{66.5}} & {\color[HTML]{009901} \textBF{83.8}} & {\color[HTML]{009901} \textBF{63.8}} \\
3.0                    & {\color[HTML]{3531FF} \textBF{83.8}} & {\color[HTML]{3531FF} \textBF{60.7}} & {\color[HTML]{3531FF} \textBF{85.6}} & {\color[HTML]{3531FF} \textBF{64.5}} & 82.4                                 & 64.7                                 & 80.2                                 & 63.6                                 & 83.0                                 & 63.3                                 \\
4.0                    & 82.4                                 & 59.4                                 & 84.5                                 & 63.8                                 & {\color[HTML]{3531FF} \textBF{84.5}} & {\color[HTML]{3531FF} \textBF{66.5}} & 80.0                                 & 63.5                                 & 82.8                                 & 63.3                                 \\
5.0                    & 82.6                                 & 59.7                                 & 83.9                                 & 63.6                                 & 84.0                                 & 66.0                                 & {\color[HTML]{009901} \textBF{82.3}} & {\color[HTML]{009901} \textBF{65.1}} & 83.2                                 & 63.6                         \\ \bottomrule[1pt]       
\end{tabular}
\end{table}

\subsubsection{Study on weighting the loss for learning motion blur robust ViTs (MBRV)}
To observe how the weight $\rho$ of the loss $\mathcal{L}_{br}$ influence the performance, we train BDTrack-DeiT with various $\rho$ values ranging from  $0.5 \times 10^{-4}$ to $5.0 \times 10^{-4}$.
The evaluation results assessed across all four datasets are shown in Table \ref{table-loss-blur}.
Obviously, BDTrack-DeiT achieves averagely the optimal performance at $\rho$ = $1.0 \times 10^{-4}$, with 84.2\% Avg. Prec. and 64.4\% Avg. Succ..
Additionally, we can observe that the second and third-best performances across these datasets are scattered both above and below the value of $1.0 \times 10^{-4}$. 
These fluctuations may be attributed to the inherent differences between these datasets.
When $\rho$ is set as $0.8 \times 10^{-4}$, BDTrack-DeiT achieves the lowest average performance, with 82.5\% Avg. Prec. and 63.1\% Avg. Succ.
This indicates that selecting an appropriate weight $\rho$ is crucial for achieving optimal tracking performance. An inappropriate 
$\rho$ can adversely affect the training process, leading to suboptimal results. The performance variations observed with different 
$\rho$ values highlight the sensitivity of the tracking algorithm to this parameter. Therefore, careful tuning of 
$\rho$ is necessary to ensure that the model performs effectively across diverse datasets and scenarios.

\subsubsection{Application to SOTA trackers}
To demonstrate the general applicability of our approach, we integrate the proposed MBRV and DEEM into two SOTA trackers: GRM \cite{gao2023generalized} and DropTrack \cite{wu2023dropmae}. Note that we substitute their original backbones with the tiny ViT, i.e., ViT-Tiny \cite{dosovitskiy2020image}, to save training time. The evaluation results on three benchmarks are presented in Table \ref{sota}. As mentioned previously, we provide only the speed of the baseline to eliminate potential nuances arising from randomness. We can observe that the inclusion of MBRV leads to noticeable improvements in both Prec. and Succ. for all baselines. Specifically, GRM and DropTrack experience increases of 1.4\% and 2.3\% in Avg. Prec., respectively, while their Avg. Succ. grows by 0.9\% and 1.6\%, respectively. 
When DEEM is further integrated, we consistently observe improvements in GPU speeds, along with only slight decreases in Prec. and Succ. More specifically, the GPU speeds of all baselines demonstrate increases of more than 17.0\%, with tracking performance dropping by less than 0.5\%. These results confirm the applicability of our method in enhancing the efficiency of existing ViT-based trackers without significantly compromising tracking performance, justifying its generality.

\begin{table}[]
\centering
\scriptsize
\setlength\tabcolsep{2.5pt} 
\caption{Application of our MBRV and DEEM to two SOTA trackers, with their original backbones replaced by ViT-Tiny~\cite{dosovitskiy2020image} to reduce training time.}
\label{sota}
\begin{tabular}{cccccccccccc}
\toprule[1pt]
\multirow{2}{*}{Tracker}   & \multirow{2}{*}{MBVR} & \multirow{2}{*}{DEEM} & \multicolumn{2}{c}{UAVDT}     & \multicolumn{2}{c}{VisDrone2018}  & \multicolumn{2}{c}{UAV123@10fps} & \multicolumn{2}{c}{Avg}       & \multirow{2}{*}{Avg.FPS} \\
                           &                       &                       & Prec.         & Succ.         & Prec.         & Succ.         & Prec.           & Succ.          & Prec.         & Succ.         &                          \\ \hline
\multirow{3}{*}{GRM~\cite{gao2023generalized}}       &                       &                       & 78.1          & 56.9          & 85.5          & 64.8          & 82.6            & 65.3           & 82.1          & 62.3          & 218.6                    \\
                           &$\checkmark$                       &                       & \textBF{80.4} & \textBF{58.4} & \textBF{86.2} & \textBF{65.3} & \textBF{84.0}   & \textBF{66.0}  & \textBF{83.5} & \textBF{63.2} & -                        \\
                           &$\checkmark$                       &$\checkmark$                       & 79.9          & 58.1          & 85.5          & 64.8          & 83.5            & 65.7           & 83.0          & 62.9          & \textBF{257.3}$_{\uparrow 17.8\%}$           \\ \hline
\multirow{3}{*}{DropTrack~\cite{wu2023dropmae}} &                       &                       & 79.4          & 57.5          & 81.9          & 61.8          & 82.5            & 65.1           & 81.3          & 61.5          & 211.4                    \\
                           &$\checkmark$                        &                      & \textBF{81.6} & \textBF{58.9} & \textBF{85.1} & \textBF{64.2} & \textBF{84.0}   & \textBF{66.1}  & \textBF{83.6} & \textBF{63.1} & -                        \\
                           &$\checkmark$                       &$\checkmark$                       & 81.0          & 58.4          & 84.7          & 63.8          & 83.7            & 65.8           & 83.1          & 62.7          & \textBF{248.6}$_{\uparrow 17.5\%}$    \\ \bottomrule[1pt]      
\end{tabular}
\end{table}

\begin{table}[h]
\scriptsize
\centering
\setlength\tabcolsep{4.5pt}
\caption{Comparison of performance and speed between our DEEM and various early exit methods on three UAV tracking datasets. Note that the baseline represents the enhanced tracker, which already includes the proposed MBVR.}\label{table:exiting_methods}
\begin{tabular}{cccccccccc}
\toprule[1pt]
                          & \multicolumn{2}{c}{UAVDT}                                                   & \multicolumn{2}{c}{VisDrone2018}                                            & \multicolumn{2}{c}{UAV123@10fps}                                            & \multicolumn{2}{c}{Avg.}                                                    &                                       \\
\multirow{-2}{*}{Tracker} & Prec.                                & Succ.                                & Prec.                                & Succ.                                & Prec.                                & Succ.                                & Prec.                                & Succ.                                & \multirow{-2}{*}{Avg.FPS}             \\ \hline
baseline                  & 84.5                                 & 61.4                                 & 86.1                                 & 65.1                                 & 84.2                                 & 66.2                                 & 84.9                                 & 64.2                                 & 228.7                                 \\
+ DEEM                    & {\color[HTML]{FE0000} \textBF{84.1}} & {\color[HTML]{FE0000} \textBF{61.0}} & {\color[HTML]{FE0000} \textBF{85.2}} & {\color[HTML]{FE0000} \textBF{64.3}} & {\color[HTML]{3531FF} \textBF{83.5}} & {\color[HTML]{3531FF} \textBF{65.9}} & {\color[HTML]{FE0000} \textBF{84.3}} & {\color[HTML]{FE0000} \textBF{63.7}} & 279.4$_{\uparrow 22.2\%}$                                 \\
+ PABEE~\cite{zhou2020bert}                   & 81.7                                 & 58.5                                 & {\color[HTML]{3531FF} \textBF{84.1}} & {\color[HTML]{3531FF} \textBF{63.8}} & 82.1                                 & 63.8                                 & 82.6                                 & 62.0                                 & {\color[HTML]{3531FF} \textBF{298.1}}$_{\uparrow 30.7\%}$ \\
+ ViT-EE~\cite{bakhtiarnia2021multi}                  & 81.1                                 & 58.2                                 & 82.5                                 & 61.7                                 & 83.0                                 & 65.6                                 & 82.2                                 & 61.8                                 & {\color[HTML]{FE0000} \textBF{318.7}}$_{\uparrow 39.4\%}$ \\
+ LGViT~\cite{Xu2023LGViTDE}                   & {\color[HTML]{3531FF} \textBF{82.7}} & {\color[HTML]{3531FF} \textBF{59.2}} & 83.1                                 & 62.0                                 & {\color[HTML]{FE0000} \textBF{84.1}} & {\color[HTML]{FE0000} \textBF{66.2}} & {\color[HTML]{3531FF} \textBF{83.3}} & {\color[HTML]{3531FF} \textBF{62.5}} & 264.5$_{\uparrow 15.7\%}$       \\ \bottomrule[1pt]                         
\end{tabular}
\end{table}

\subsubsection{Comparison with Existing Early Exit Methods}

We have conducted a comparative experiments between our method and existing early exit mechanisms, including PABEE~\cite{zhou2020bert}, ViT-EE~\cite{bakhtiarnia2021multi}, and LGViT~\cite{Xu2023LGViTDE}. The evaluation results are shown in Table~\ref{table:exiting_methods}. As observed, while these methods employ more complex architectures, they introduce additional complexity, struggle to balance speed and accuracy, and may exit prematurely, resulting in suboptimal performance in challenging tracking scenarios. Specifically, although PABEE and ViT-EE achieve faster speeds than our DEEM, their performance is inferior, with an average drop of over 2.0\% in precision and success rate. Additionally, while LGViT achieves relatively better performance than PABEE and ViT-EE, it comes at the cost of even slower speeds. In contrast, with the inclusion of our DEEM, the baseline's performance drops by less than 0.6\% in average precision and success rate, while achieving a speedup of over 22\%. These comparisons highlight the effectiveness of the proposed DEEM in UAV tracking.

\begin{figure}[h]
	\centering
\includegraphics[width=0.65\textwidth]{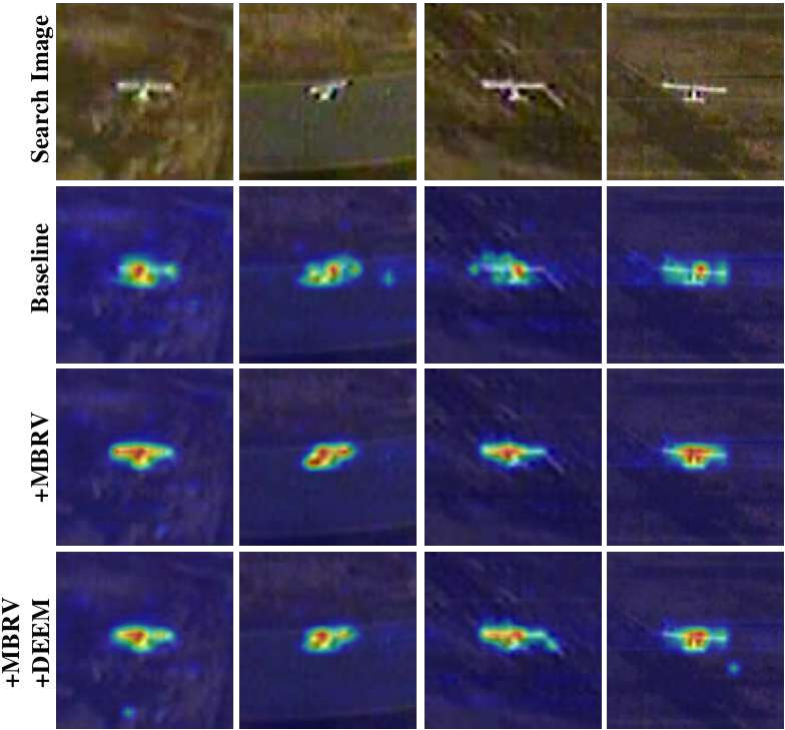}
\caption{Visualization of attention maps. The first row displays the original search images, while the second, third, and fourth rows showcase the attention maps generated by BDTrack-DeiT without and with incorporating the proposed components. }
 \label{fig_vis_attn}
\end{figure}

\subsubsection{Qualitative Results}
\begin{figure*}[h]
	\centering
\includegraphics[width=0.65\textwidth]{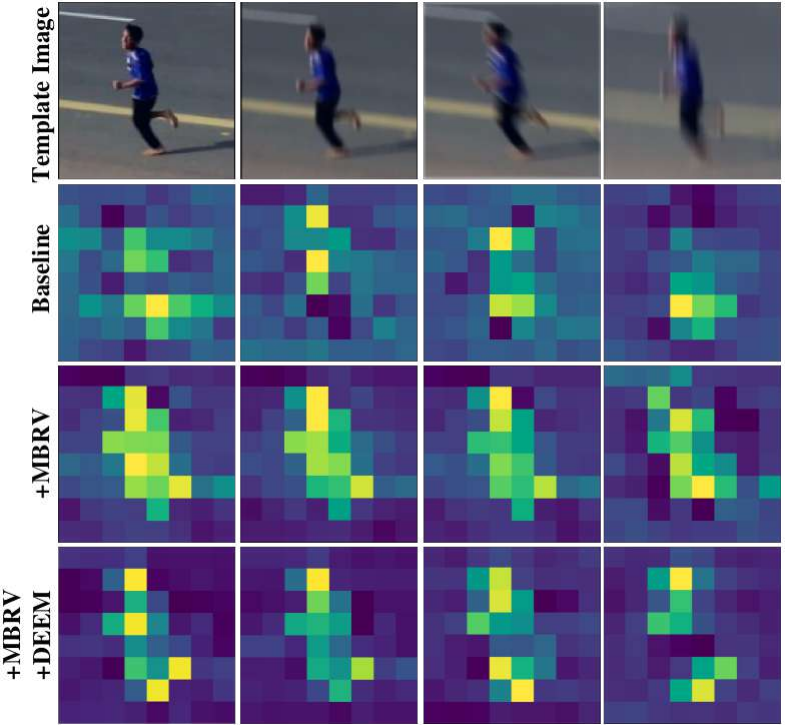}
    \caption{Visualization of feature maps. For each group, we present the original template images and their blurred versions (top row), followed by the feature maps generated by BDTrack-DeiT without (second row) and with (third and fourth rows) the proposed components. 
    The the kernel size for linear motion blur is 3, 5 , 7, from left to right.}
 \label{fig_vis_feat}
\end{figure*}

Our BDTrack improves ViT robustness to motion blur by enforcing target feature invariance.
To illustrate the effectiveness of our components, Fig.\ref{fig_vis_attn} and Fig.\ref{fig_vis_feat} show example attention and feature maps generated by BDTrack with and without the proposed modules (i.e., MBRV and DEEM), under motion blur conditions.
For clarity, we use suffixes to indicate model variants:
‘*’ denotes the baseline without our components,
‘+’ adds MBRV to the baseline,
and ‘++’ includes both MBRV and DEEM.

\textit{Example attention maps of the search images.} 
In Fig. \ref{fig_vis_attn}, the first row shows the original search images from an example sequence (i.e., uav1 from UAV123 \cite{mueller2016benchmark}).
The images in the second, third, and fourth rows are the corresponding attention maps generated by BDTrack-DeiT*, BDTrack-DeiT+, and BDTrack-DeiT++ (i.e., BDTrack-DeiT), respectively.
As can be seen, incorporating the MBRV component (i.e., BDTrack-DeiT+) and both the MBRV and DEEM components (i.e., BDTrack-DeiT++) yields very similar results. Specifically, their visual focus on the targets is more pronounced compared to the baseline tracker (i.e., BDTrack-DeiT*). This demonstrates that the MBRV component is able to enhance the tracker's ability to concentrate on the target, especially under motion blur conditions, thereby improving overall tracking performance.

\textit{Example feature maps of the template images.} 
In each group of Fig. \ref{fig_vis_feat}, the first row displays the original template and its blurred versions using the linear motion blur method \cite{Dwibedi2017CutPA} with kernel sizes 3, 5, 7 from left to right, respectively. 
The corresponding feature maps produced by BDTrack-DeiT*, BDTrack-DeiT+, and BDTrack-DeiT++ are shown in the second, third, and fourth rows, respectively.
This comparison provides a visual representation of the impact of the proposed MBRV and DEEM components. 
As can be seen, the feature maps generated by BDTrack-DeiT+ and BDTrack-DeiT++ show more consistency with changes in kernel size, whereas the feature maps from BDTrack-DeiT* display more significant variations, especially at larger kernel sizes.
These qualitative results provide visual evidence for the effectiveness of our method in learning motion blur robust feature representations with ViTs.

\begin{figure}[t]
\centering
\includegraphics[width=0.65\textwidth,height=0.35\textwidth]{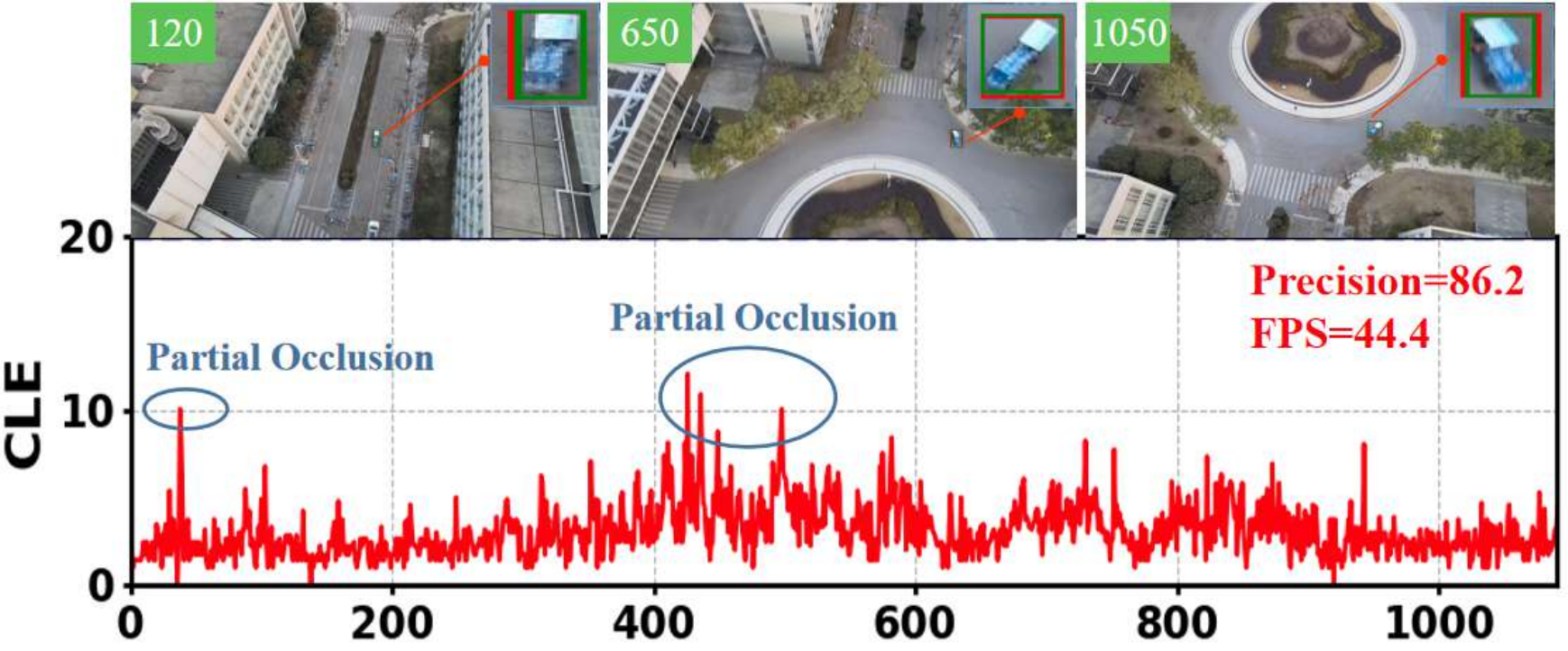}

\vskip 1ex 
\includegraphics[width=0.65\textwidth,height=0.35\textwidth]{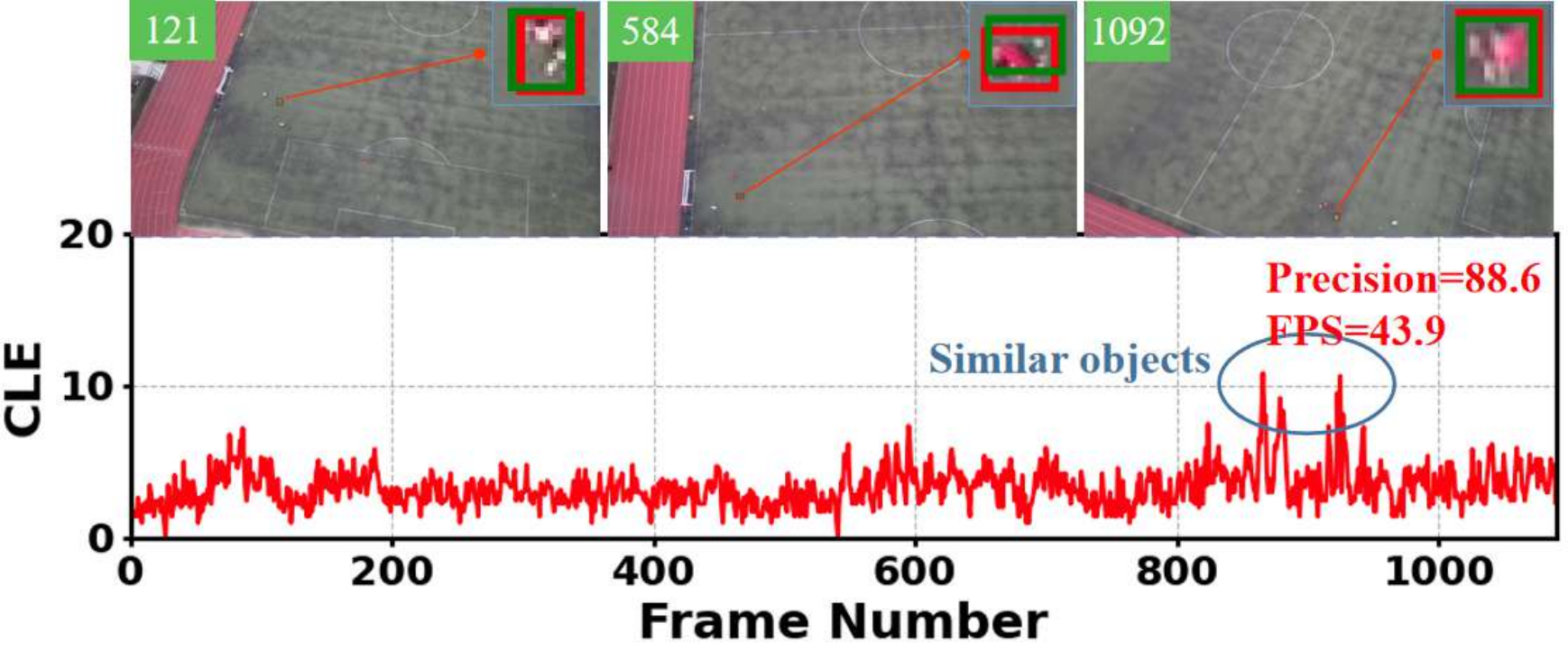}
\caption{Real-world test on an embedded device. The CLE plots illustrate the frame-wise performance, where errors below a CLE of 20 pixels are usually considered acceptable. Note that the ground truth and tracking results are denoted by red and green boxes, respectively.}
\label{real_world_test_fig}
\end{figure}

\subsection{Real-World Test}
To evaluate our method's tracking performance under real-world conditions, we conducted tests on the NVIDIA Jetson AGX Xavier 32GB, a platform well-suited for UAV applications due to its compactness and computational efficiency. We used the UAVTrack112\_L dataset \cite{fu2021onboard} to evaluate our BDTrack-DeiT, which cover a variety of UAV scenes. Fig. \ref{real_world_test_fig} showcases two diverse real-world scenarios where the objects are subjected to motion blur, low resolution, fast motion challenges among others. As can be seen, our BDTrack-DeiT demonstrates robust and satisfactory performance under these challenging real-world conditions. Specifically, it maintains a center location error (CLE) of fewer than 20 pixels in all test frames, indicating high precision in tracking. Furthermore, BDTrack-DeiT consistently achieves real-time average speeds of 44 FPS. These results highlight our tracker’s robustness and efficiency, making it well-suited for UAV applications that demand high performance and speed.

\begin{figure}[t]
\centering
\includegraphics[width=0.55\textwidth]{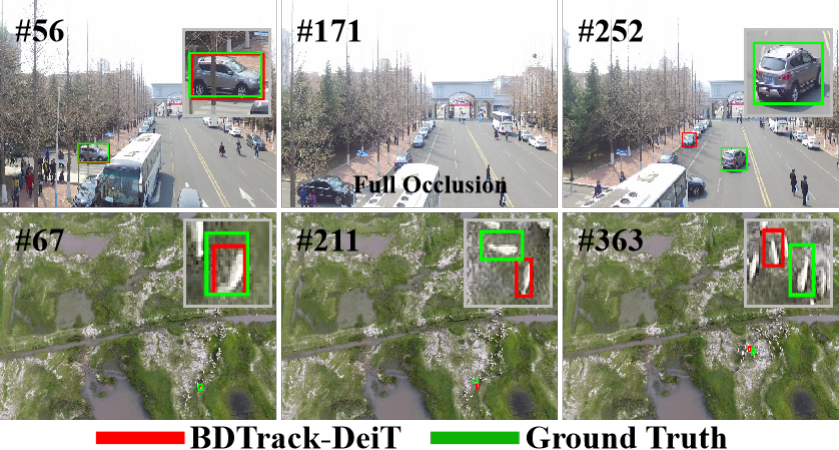}
\caption{BDTrack limitations in complex tracking situations: Tracking errors occur when the object is fully occluded, indistinguishable from the surroundings, or occupies a very small region.}
\label{fig_fail_case}
\end{figure}

\section{Limitations and future work}
\label{section_limitations}

\subsection{Limitations}

We present several representative failure cases in Fig. \ref{fig_fail_case}.
Although our BDTrack demonstrates strong performance in both efficiency and motion blur robustness, it still faces challenges in certain complex tracking scenarios. Specifically, the current design lacks an explicit mechanism for recovering target features after full occlusion, which may lead to identity loss or tracking failure once the target reappears (as shown in the top row). Moreover, the tracker’s performance degrades when dealing with extremely small or visually similar (as shown in the bottom row). This is primarily due to the limited spatial resolution of high-level features and the absence of fine-grained contrast enhancement, which affects the ability to distinguish the target from the background and increases the risk of drift.

\subsection{Future work}

To address these limitations, future work will focus on enhancing the tracker’s robustness in scenarios involving full occlusion and low visibility. One potential direction is to incorporate lightweight memory mechanisms that enable the tracker to re-identify the target after temporary disappearance. Additionally, we aim to improve the model’s sensitivity to fine-grained visual cues by integrating high-resolution refinement modules, thereby facilitating more accurate localization of small targets.

\section{Conclusions}\label{section_conclusion}

In this work, we presented a unified framework that explores both efficiency and motion blur robustness for real-time UAV tracking with Vision Transformers (ViTs). Our design introduces an adaptive computation paradigm via a dynamic early exiting mechanism, significantly reducing inference cost without compromising accuracy. Furthermore, we proposed a motion blur robust learning strategy that enforces feature invariance between clear and blurred templates, enhancing the tracker's robustness under adverse conditions.

Extensive experiments conducted on four UAV tracking benchmarks demonstrate the effectiveness and versatility of our approach. In particular, BDTrack-DeiT achieves a more favorable speed-accuracy trade-off compared to existing state-of-the-art trackers. It outperforms lightweight models such as Aba-ViTrack~\cite{li2023adaptive}, running approximately 1.6 times faster on GPU and 1.3 times faster on CPU, and surpasses deeper models like ROMTrack~\cite{cai2023robust} and EVPTrack~\cite{shi2024evptrack} with speed improvements of about 5 and 11 times, respectively, while achieving comparable or even better accuracy.
The experimental results demonstrate that our BDTrack-DeiT achieves state-of-the-art performance in real-time UAV tracking.

\bibliographystyle{IEEEtran}
\bibliography{main}




\end{document}